\def\year{2021}\relax
\documentclass[letterpaper]{article} %
\usepackage{aaai21}  %
\usepackage{times}  %
\usepackage{helvet} %
\usepackage{courier}  %
\usepackage[hyphens]{url}  %
\usepackage{graphicx} %
\def\UrlFont{\rm}  %
\usepackage{natbib}  %
\usepackage{caption} %
\newcommand{\bx}{\mathbf{x}}

\newcommand{\bo}{\mathbf{o}}
\newcommand{\br}{\mathbf{r}}
\newcommand{\be}{\mathbf{e}}
\newcommand{\bh}{\mathbf{h}}

\newcommand{\bc}{\mathbf{c}}

\usepackage{amsmath}
\usepackage{amssymb}
\usepackage{bbm}
\usepackage{enumitem}
\usepackage{booktabs}

\usepackage{subcaption}
\usepackage{todonotes}

\usepackage[nameinlink]{cleveref}

\crefname{section}{Section}{Sections}
\crefname{subsection}{Section}{Sections}
\crefname{subsubsection}{Section}{Sections}
\crefname{figure}{Figure}{Figures}
\crefname{table}{Table}{Tables}
\crefname{subfigure}{Figure}{Figures}
\crefname{equation}{Equation}{Equations}
\crefname{appendix}{Appendix}{Appendix}

\usepackage{titlesec}

\title{Hierarchical Relational Inference }
\author{
    Aleksandar Stani{\'c}, Sjoerd van Steenkiste, J\"{u}rgen Schmidhuber%
    \\
}

\affiliations{
    Swiss AI Lab IDSIA, USI, SUPSI \\
    Lugano, Switzerland \\

    \texttt{\{aleksandar, sjoerd, juergen\}@idsia.ch} \\

}

\begin{document}

\maketitle

\begin{abstract}
Common-sense physical reasoning in the real world requires learning about the interactions of objects and their dynamics.
The notion of an abstract object, however, encompasses a wide variety of physical objects that differ greatly in terms of the complex behaviors they support. 
To address this, we propose a novel approach to physical reasoning that models objects as hierarchies of parts that may locally behave separately, but also act more globally as a single whole. 
Unlike prior approaches, our method learns in an unsupervised fashion directly from raw visual images to discover objects, parts, and their relations. 
It explicitly distinguishes multiple levels of abstraction and improves over a strong baseline at modeling synthetic and real-world videos. %

\end{abstract}

\section{Introduction}

Common-sense physical reasoning in the real world involves making predictions from complex high-dimensional observations.
Humans somehow discover and represent abstract objects to compactly describe complex visual scenes in terms of `building blocks' that can be processed separately~\cite{spelke2007core}.
They model the world  by reasoning about dynamics of high-level objects such as footballs and football players and the consequences of their interactions.
It is natural to expect that artificial agents operating in the real world will benefit from a similar approach~\cite{lake2015human,greff2020binding}.

Real world objects  vary greatly in terms of their properties. This complicates modelling their dynamics. 
Some have deformable shapes, e.g., clothes, or consist of parts that support a variety of complex behaviors, e.g., arms and fingers of a human body.
Many objects can be viewed as a \emph{hierarchy} of parts that locally behave somewhat independently, but also act more globally as a single whole~\cite{mrowca2018flexible,lingelbach2020towards}. 
This suggests to simplify models of object dynamics by explicitly distinguishing multiple levels of abstraction, separating hierarchical sources of influence.

Prior approaches to common-sense physical reasoning explicitly consider objects and relations at a representational level, e.g.,~\cite{chang2016compositional,battaglia2016interaction,van2018relational,kipf2018neural}.
They decompose complex physical interactions in the environment into pairwise interactions between objects, modelled efficiently by Graph Networks~\cite{battaglia2018relational}.
Here the representation of each object is updated at each time step by propagating `messages' through the corresponding interaction graph.
While recent approaches (specifically) address the challenge of learning object representations from raw visual data~\cite{greff2017neural,kosiorek2018sequential,van2018relational,burgess2019monet,greff2019multi} and of dynamically inferring relationships between objects~\cite{van2018relational,kipf2018neural,goyal2019recurrent,veerapaneni2019entity}, reasoning about the dynamics and interactions of complex objects remains difficult without incorporating additional structure.
On the other hand, approaches that consider part-based representations of objects and hierarchical interaction graphs lack the capacity to learn from raw images and dynamically infer relationships~\cite{mrowca2018flexible, lingelbach2020towards}.

\begin{figure}[t]
    \centering
    \includegraphics[width=\linewidth]{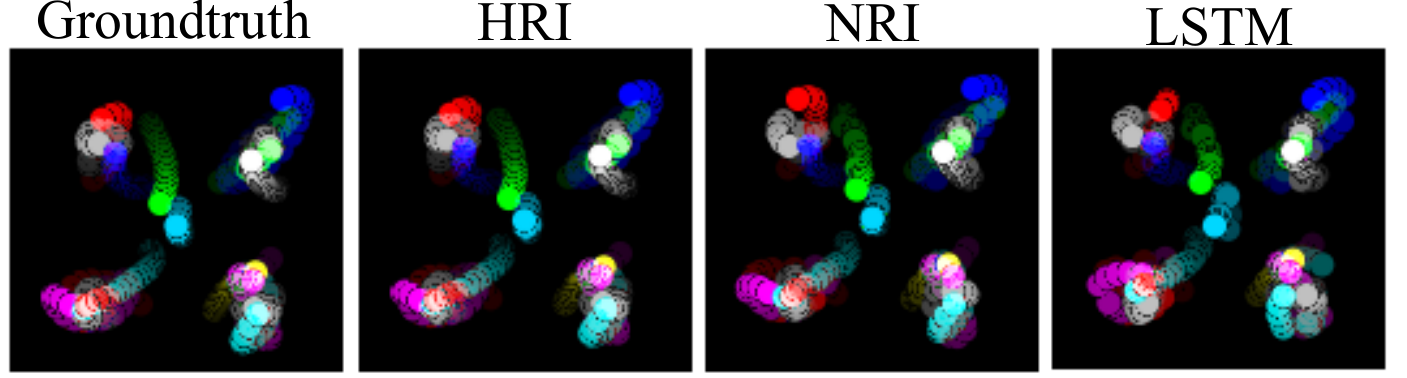}
\caption{ 
HRI outperforms baselines at modeling interacting objects that are coupled via hierarchically organized springs.
} 
\label{fig:traj-traces}
\end{figure}%

\begin{figure*}[t]
    \centering
    \includegraphics[width=0.8\textwidth]{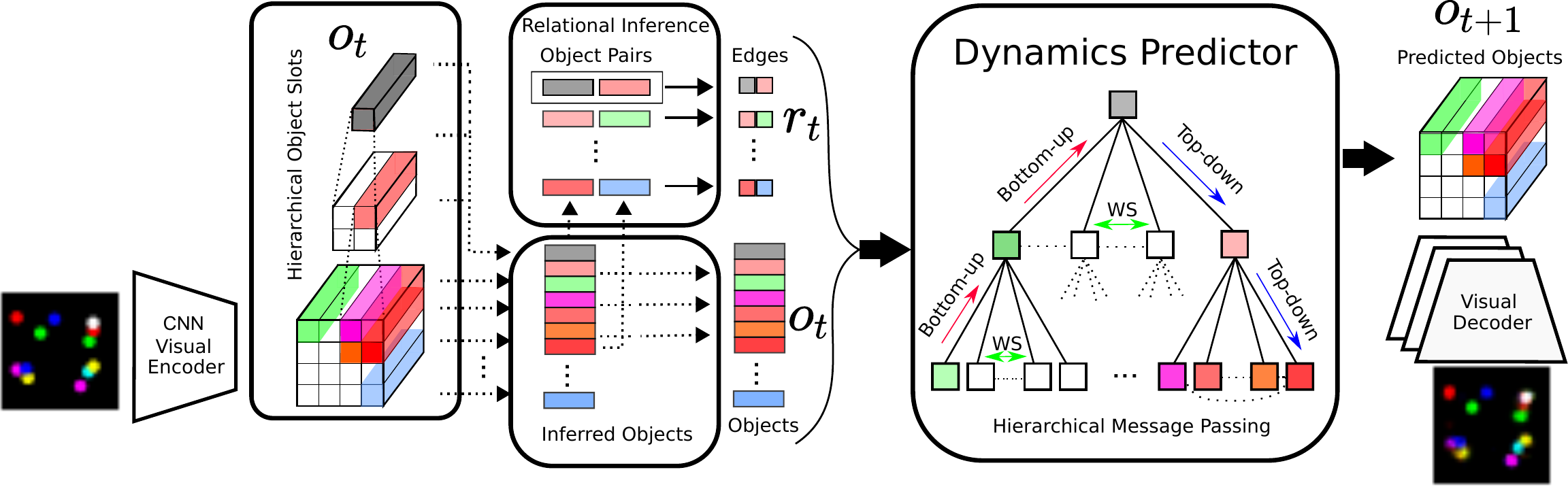}
    \caption{The proposed HRI model. 
    An encoder infers part-based object representations, which are fed to a relational inference module to obtain a hierarchical interaction graph.
    A dynamics predictor uses hierarchical message-passing to make predictions about future object states.
    Their `rendering', produced by a decoder, is compared to the next frame to train the system.
    }
    \label{fig:hri}
\end{figure*}

Here we propose \emph{Hierarchical Relational Inference} (HRI), a novel approach to common-sense physical reasoning capable of learning to discover objects, parts, and their relations, directly from raw visual images in an unsupervised fashion.
HRI extends \emph{Neural Relational Inference} (NRI)~\cite{kipf2018neural}, which infers relations between objects and models their dynamics while assuming access to their state (e.g., obtained from a physics simulator).
HRI improves upon NRI in two regards.
Firstly, it considers part-based representations of objects and infers \emph{hierarchical} interaction graphs to simplify modeling the dynamics (and interactions) of more complex objects.
This necessitates a more efficient message-passing approach that leverages the hierarchical structure, which we will also introduce.
Secondly, it provides a mechanism for applying NRI (and thereby HRI) to raw visual images that infers part-based object representations spanning multiple levels of abstraction.

\textbf{Our main contributions} are as follows: 
(i) We introduce HRI, an end-to-end approach for learning hierarchical object representations and their relations directly from raw visual input. 
(ii) It includes novel modules for extracting a hierarchical part-based object representations and for hierarchical message passing. 
The latter can operate on a (hierarchical) interaction graph more efficiently by propagating effects between all nodes in the graph in a single message-passing phase.
(iii) On a trajectory prediction task from object states, we demonstrate that the hierarchical message passing module is able to discover the latent hierarchical graph and greatly outperforms strong baselines (\cref{fig:traj-traces}). 
(iv) We also demonstrate how HRI is able to infer objects and relations directly from raw images.
(v) We apply HRI to synthetic and real-world physical prediction tasks, including real-world videos of moving human bodies, and demonstrate improvements over strong baselines.

\section{Method}

Motivated by how humans learn to perform common-sense physical reasoning, we propose Hierarchical Relational Inference (HRI).
It consists of a \emph{visual encoder}, a \emph{relational inference module}, a \emph{dynamics predictor}, and a \emph{visual decoder}. All are trained end-to-end in an unsupervised manner.
First, the visual encoder produces hierarchical (i.e. part-based) representations of objects that are grounded in the input image.
This representation serves as input to the relational inference module, which infers pairwise relationships between objects (and parts) given by the edges in the corresponding interaction graph.
The dynamics predictor performs hierarchical message-passing on this graph, using the learned  representations of parts and objects for the nodes.
The resulting predictions (based on the updated representations) are then decoded back to image space using the visual decoder, to facilitate an unsupervised training objective. 
An overview is shown in Figure~\ref{fig:hri}.
We note that HRI consists of standard building blocks (CNNs, RNNs, and GNNs) that are well understood.
In this way, we add only a minimal inductive bias, which helps facilitate scaling to more complex real-world visual settings, as we will demonstrate.

\subsection{Inferring Objects, Parts, and their Relations}

To make physical predictions about a stream of complex visual observations, we will focus on the underlying \emph{interaction graph}.
It distinguishes objects or parts (corresponding to nodes) and the relations that determine interactions between them (corresponding to the edges), which must be inferred.
Using this more abstract (neuro-symbolic) representation of a visual scene allows us to explicitly consider certain invariances when making predictions (e.g., the number of objects present) and reasoning about complex interactions in terms of simpler pair-wise interactions between objects.

\paragraph{Inferring Object/Part Representations}

The task of the visual encoder is to infer separate representations for each object from the input image.
Intuitively, these representations contain information about its state, i.e., its position, behavior and appearance.
In order to relate and compare these representations efficiently, it is important that they are described in a common format.
Moreover, since we are concerned with a hierarchical (i.e. part-based) representation of objects, we also require a mechanism to relate the part representations to the corresponding object representation. 

Here we address these challenges by partitioning the feature maps learned by a CNN according to their spatial coordinates to obtain object representations.
This is a natural choice, since CNNs are known to excel at representation learning for images~\cite{ciregan2012multi, krizhevsky2012imagenet} and because weight-sharing across spatial locations ensures that the resulting object representations are described in a common format.
Indeed, several others have proposed to learn object representations in this way~\cite{santoro2017simple,zambaldi2018deep}. 
Here, we take this insight a step further and propose to learn hierarchical object representations in a similar way.
In particular, we leverage the insight that the parts belonging to real-world objects tend to be spatially close, to apply a sequence of convolutions followed by down-sampling operations to extract object-level representations from part-level representations (left side of Figure~\ref{fig:hri}).
While this leaves the network with ample freedom to develop its own internal notion of an object, we find that representational slots learn to describe physical objects (\cref{fig:imgs-nll-4-3-slots}).

The choice of kernel-size and degree of down-sampling allow us to adjust how representations at one level of abstraction are combined at the next level.
Similarly, the  parameters of the CNN layers that produce the initial set of feature maps determine the size of the input region captured by these `spatial slots'~\cite{greff2020binding} at the lowest level of abstraction.
Note the distinction between \emph{parts}, \emph{objects} and \emph{slots}.
Parts refer to objects at the lowest level of the visual hierarchy, while the more general notion of an object applies to nodes at all levels. 
Slots are variable placeholders (of a function) at a representational level, which at each point in time are expected to contain information about a particular object/part. 
Therefore, an architectural hierarchy of slots reflects a hierarchy of objects.
In general, we construct a 3-level part-based hierarchy, which is then fed into the \emph{relational module}. 

\paragraph{Neural Relational Inference}

To infer relations between object representations, we will make use of NRI~\cite{kipf2018neural}, which learns \emph{explicit} relations.
This is advantageous, since it allows one to incorporate prior beliefs about the overall connectivity of the interaction graph (e.g., a degree of sparsity) and associate a representation with each relation to distinguish between multiple different relation types.
By default, NRI takes as input a set of object trajectories (states) and infers their pairwise relations (edges) using a Graph Neural Network (GNN)~\cite{scarselli2009, gilmer2017neural, battaglia2018relational}.
It assumes a static interaction graph, and performs relational inference by processing the entire input sequence at once, i.e., the whole sequence (length 50) of a particular object is concatenated, and only a single, static, ``node embedding'' is created via an MLP.
In contrast, we will consider a dynamic interaction graph, since objects move across the image and may end up in different spatial slots throughout the sequence.
This is achieved by inferring edges at each time step based on the ten most recent object states, concatenating the latent vectors of a particular object and using an MLP to obtain a ``node embedding''. 

More formally, given a graph $\mathcal{G}=(\mathcal{V},\mathcal{E})$ with nodes (objects) $o\in\mathcal{V}$ and edges (relations) $r_{i,j}=(o_i,o_j)\in\mathcal{E}$, NRI defines a single node-to-node message passing operation in a GNN similar to~\cite{gilmer2017neural}:
\begin{equation}
\be_{i,j} = f_{e}([\bo_i,\bo_j,\br_{i,j}]), \
\bo_{j}' = f_{o}([\textstyle\sum_{i\in \mathcal{N}_{\bo_j}} \be_{i,j},\bo_{j}]) \label{eq:gnn_mp}
\end{equation}
where $\be_{i,j}$ is an embedding (effect) of the relation $\br_{i,j}$ between objects $\bo_i$ and $\bo_j$, $\bo_{j}'$ is the updated object embedding, $\mathcal{N}_{j}$ the set of indices of nodes connected by an incoming edge to object $\bo_j$ and $[\cdot,\cdot]$ indicates concatenation.
Functions $f_{o}$ and $f_{e}$ are node- and edge-specific neural networks (MLPs in practice).
By repeatedly applying \eqref{eq:gnn_mp}, multiple rounds of message passing can be performed.

The NRI `encoder' receives as input a sequence of object state trajectories $\bo=(\bo^1,...,\bo^T)$, which in our case are inferred. 
It consists of a GNN $f_{\phi}$ that defines a probability distribution over edges $q_{\phi}(\br^t_{ij}|\bo^{t-k:t})=\mathrm{softmax}(f_{\phi}(\bo^{t-k:t})_{ij})$, where $k$ is the window size, and relations are one-hot encoded.
The GNN performs the following message passing operations, where the initial node representations $\bo_i$ are obtained by concatenating the corresponding object states across the window:
\begin{align*}
& \bo_{j}' = f_{\mathrm{o}}^1(\bo_j), \
\be_{i,j}' = f_{e}^1([\bo_i',\bo_j']), \
\bo_{j}'' = f_{o}^2(\textstyle\sum_{i\neq j}\be_{i,j}'), \\
& \be_{i,j}'' = f_{e}^2([\bo_i'',\bo_j'']), \
f_{\phi}(\bo^{t-k:t})_{ij} = \be_{i,j}''
\end{align*}
where $\phi$ contains the parameters of the message-passing functions, which are simple MLPs, and $o', e'$ and $o'', e''$ are node- and edge-embeddings after first and second message passing operations respectively.
To backpropagate through the sampling from $q_{\phi}(\br_{ij}|\bo)$, NRI uses a continuous approximation of the discrete distribution to obtain gradients via the reparameterization trick~\cite{maddison2016concrete, jang2016categorical}. 

\subsection{Physical Reasoning}

Physical reasoning is performed by the \emph{dynamics predictor}, which leverages the inferred object representations and edges to predict object states at the next time step.
To distinguish between representations at different levels of abstractions, HRI makes use of \emph{hierarchical message passing}.
We will also use recurrent units in the non-Markovian setting.

\paragraph{Hierarchical message passing}

The default approach to physical reasoning based on message-passing employed in NRI can only propagate effects between directly connected nodes.
This is costly, as it requires several iterations for information to propagate across the whole graph, especially when the number of nodes increases (a consequence of modeling objects as hierarchies of parts).
Alternatively, we can leverage the hierarchical structure of the interaction graph to propagate all effects across the entire graph in a \emph{single} step, i.e. evaluating each relation only once.
To achieve this, we introduce a hierarchical message passing module which propagates effects between objects using a three-phase sequential mechanism, loosely inspired by \citeauthor{mrowca2018flexible}.

Starting from the leaf nodes, the \emph{bottom-up phase} computes the effect on parent nodes $\bo_p$ based on messages from its children,
$\be_p^1 = \be_p^0 + f^{bu}_{MP}(\{\be_c^0\}_{c\in\mathcal{C}_p}, \be_p^0, \{\br_{cp}\}_{c\in\mathcal{C}_p})$ 
where $\mathcal{C}_{p}$ is the set of children indices of object $\bo_p$ and the initial effects $\be^0$ are simply the object embeddings.
In this way, global information is propagated from every node in the hierarchy to the root node.
Afterwards, the bottom-up effect $\be_i^1$ on node $\bo_i$ is combined with effects from its siblings (\emph{within-sibling} phase)
$\be_i^2 = \be_i^1 + f^{ws}_{MP}(\{\be_s^1\}_{s\in\mathcal{S}_i}, \be_i^1, \{\br_{si}\}_{s\in\mathcal{C}_i})$,
where $\mathcal{S}_{i}$ is the set of sibling indices of object $\bo_i$.
Starting from the root node, the \emph{top-down} phase then propagates top-down effects that are incorporated by computing $\be_c^3 = \be_c^2 + f^{td}_{MP}(\be_p^2, \be_c^2, \br_{pc})$ for all children $\bo_c$ based on its parent $\bo_p$.
Functions $f^{bu}_{MP}, f^{ws}_{MP},$ and $f^{td}_{MP}$ perform a single node-to-edge and edge-to-node message passing operation as in \eqref{eq:gnn_mp} and have shared weights.
Note that this mechanism is independent of the choice of object and relational inference module and can act on any hierarchical interaction graph.

\paragraph{Dynamics predictor}

Physical reasoning is performed by the dynamics predictor, which predicts future object states 
$p_\theta(\bo^{t+1}|\bo^{1:t},\br^{1:t})$
from the sequence of object states and interactions.
We implement this as in the NRI `decoder'~\cite{kipf2018neural}, i.e. using a GNN that passes messages between objects, but with two notable differences.
Firstly, we will pass messages \emph{only} if an edge is inferred between two nodes, as opposed to also considering a separate dynamics predictor for the ``non-edge'' relation that causes information exchange between unconnected nodes\footnote{Although messages are passed only if an edge between two nodes exists, a ``non-edge'' categorical variable is used to allow the model to infer that there is no edge between two nodes.}.
Secondly, we will leverage the hierarchical structure of the inferred interaction graph to perform \emph{hierarchical message-passing}.

If we assume Markovian dynamics, then we have $p_\theta(\bo^{t+1}|\bo^{1:t},\br^{1:t})=p_\theta(\bo^{t+1}|\bo^t,\br^t)$ and can use hierarchical message passing to predict object states at the next step:
\begin{align}
p_\theta(\bo^{t+1}|\bo^t,\br^t) = \mathcal{N} (\bo^t + \Delta \bo^t,\sigma^2 \mathbf{I}),
\label{eq:ff-dyn}
\end{align}
where $\sigma^2$ is a fixed variance, $\Delta \bo^t  =  f_O([\bo^t, \be^t])$, $\be^t = f_{H}(\bo^t, \br^t)$ is the effect computed by the hierarchical message passing module $f_{H}$, and $f_O$ is an output MLP.
Notice how we learn to predict the \emph{change} in the state of an object, which is generally expected to be easier.
When encoding individual images, no velocity information can be inferred to form the object state.
In this non-Markovian case we adapt \eqref{eq:ff-dyn} to include an LSTM~\cite{hochreiter1997long} that models $p_\theta(\bo^{t+1}|\bo^{1:t},\br^{1:t})$ directly:  
\begin{align*}
& \bh^{t+1}, \bc^{t+1}  =  f_{LSTM}(\bo^t, \be^t, \bc^t), \quad
\bo^{t+1}  = f_O(\bh^{t+1}), \quad \\
& p(\bo_j^{t+1}|\bo^{1:t},\br^{1:t}) = \mathcal{N} (\bo^{t+1},\sigma^2 \mathbf{I}),
\end{align*}
where $\bc$ and $\bh$ are LSTM's cell and hidden state respectively, and $\be^t = f_{H}(\bh^t, \br^t)$.

\subsection{Learning}

Standard approaches to modelling physical interactions that do not assume access to states uses a prediction objective in pixel space~\cite{van2018relational,veerapaneni2019entity}.
This necessitates a mechanism to `render' the updated object representations.
In this case, HRI can be viewed as a type of Variational Auto-Encoder \cite{kingma2013auto, rezende2014stochastic}, where the inferred edges and objects are treated as latent variables, and the ELBO can be maximized for the predicted frames:
\begin{align}
\begin{split}
\mathcal{L}=&\mathbb{E}_{q_{\phi}(\br|\bx)}[\log p_\theta(\bx|\br, \bo)] 
- D_{\mathrm{KL}}[q_{\phi_o}(\bo|\bx)||p_{\theta_o}(\bo)] \\ 
& - D_{\mathrm{KL}}[q_{\phi_r}(\br|\bx, \bo )||p_{\theta_r}(\br)].
\end{split}
\label{eq:elbo}
\end{align}
The relational module $q_{\phi_r}(\br|\bx,\bo)$ outputs a factorized distribution over $\br_{ij}$, which in our case is a categorical variable that can take on two values (one-hot encoded) that indicate the presence of an edge between $\bo_i$ and $\bo_j$.
The edge prior $p_{\theta_r}(\br)=\prod_{i\neq j}p_{\theta_r}(\br_{ij})$ is a factorized uniform distribution, which controls the sparsity of the learned graph.
The object inference module $q_{\phi_o}(\bo|\bx)$ outputs a factorized distribution over $\bo_{i}$, and the object prior $p_{\theta_o}(\bo)$ is Gaussian, as in a standard VAE.
Given the inferred interaction graph, the dynamics predictor and visual decoder define $p_\theta(\bx|\br, \bo)$.

\paragraph{Visual Decoder}

The visual decoder renders the updated object states and we will consider two different implementations of this mapping.
The first variant, which we call \textit{SlotDec}, ensures compositionality in pixel space by decoding objects separately followed by a summation to produce the final image.
In Figure \ref{fig:hri} it is depicted as a set since individual object slots that are decoded separately, and decoders share weights.
This implements a stronger inductive bias that encourages each slot to correspond to a particular object (since images are composed of objects) and also makes it easy to inspect the representational content of each slot.
On the other hand, summation in pixel space is problematic when objects in scenes are more cluttered and occlude one another.
For this reason we implement a second variant, which we call \textit{ParDec}, where all states are decoded together as in a standard convolutional decoder.
As a result of decoding all object slots together \emph{ParDec} is less interpretable than \emph{SlotDec}, but computationally more efficient and potentially more scalable to real-world datasets since it does not make strong assumptions on how information about objects should be combined.
This may also make it easier to handle background, although this is not explored.

\begin{figure*}[t]
\centering
\begin{subfigure}[b]{0.27\textwidth}
    \centering
    \includegraphics[width=\textwidth]{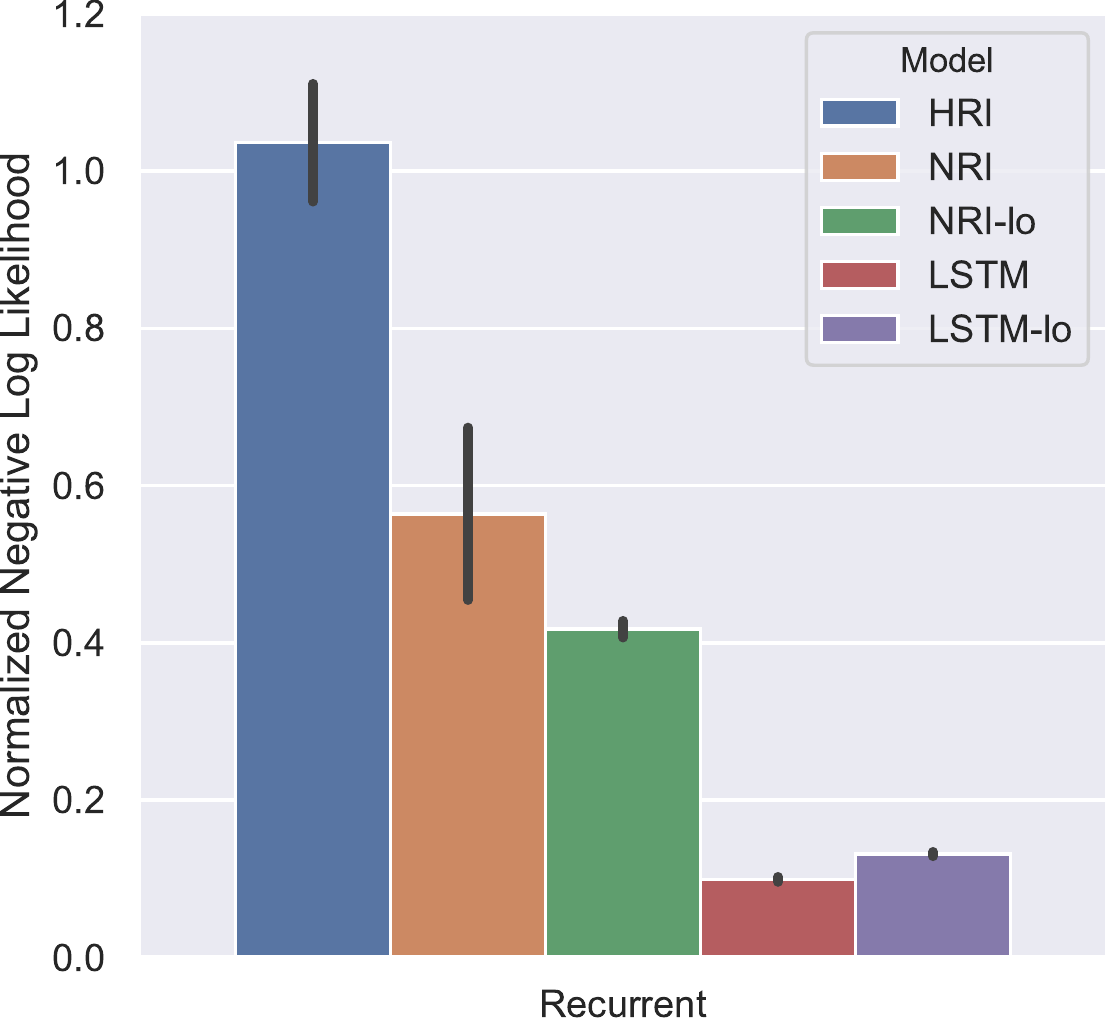}
    \caption{}
    \label{fig:traj-4-3-baselines-norm-nll}
\end{subfigure}
\hspace{0.04\textwidth}
\begin{subfigure}[b]{0.27\textwidth}   
    \centering 
    \includegraphics[width=\textwidth]{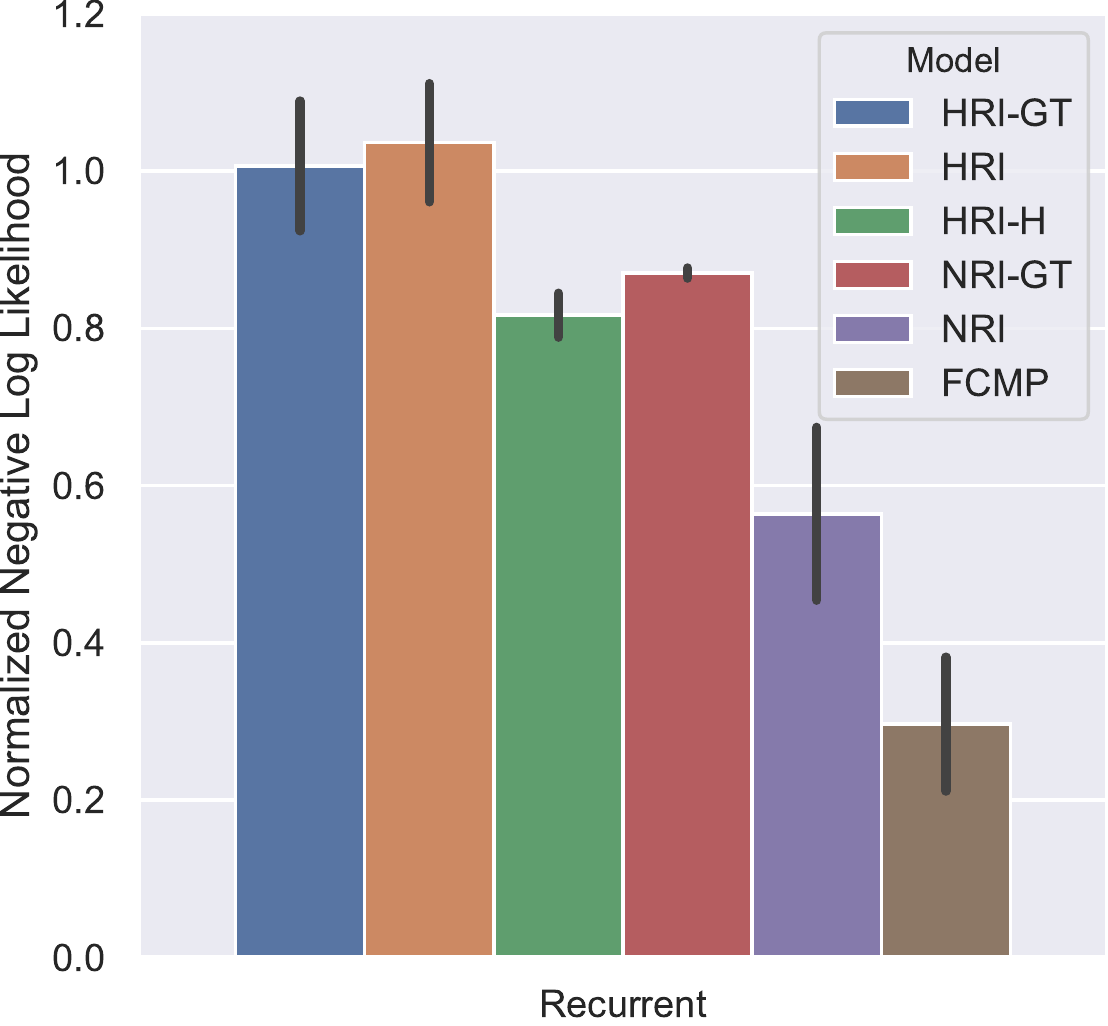}
    \caption{}
    \label{fig:traj-4-3-ablation-norm-nll}
\end{subfigure}
\hspace{0.04\textwidth}
\begin{subfigure}[b]{0.27\textwidth}   
    \centering 
    \includegraphics[width=\textwidth]{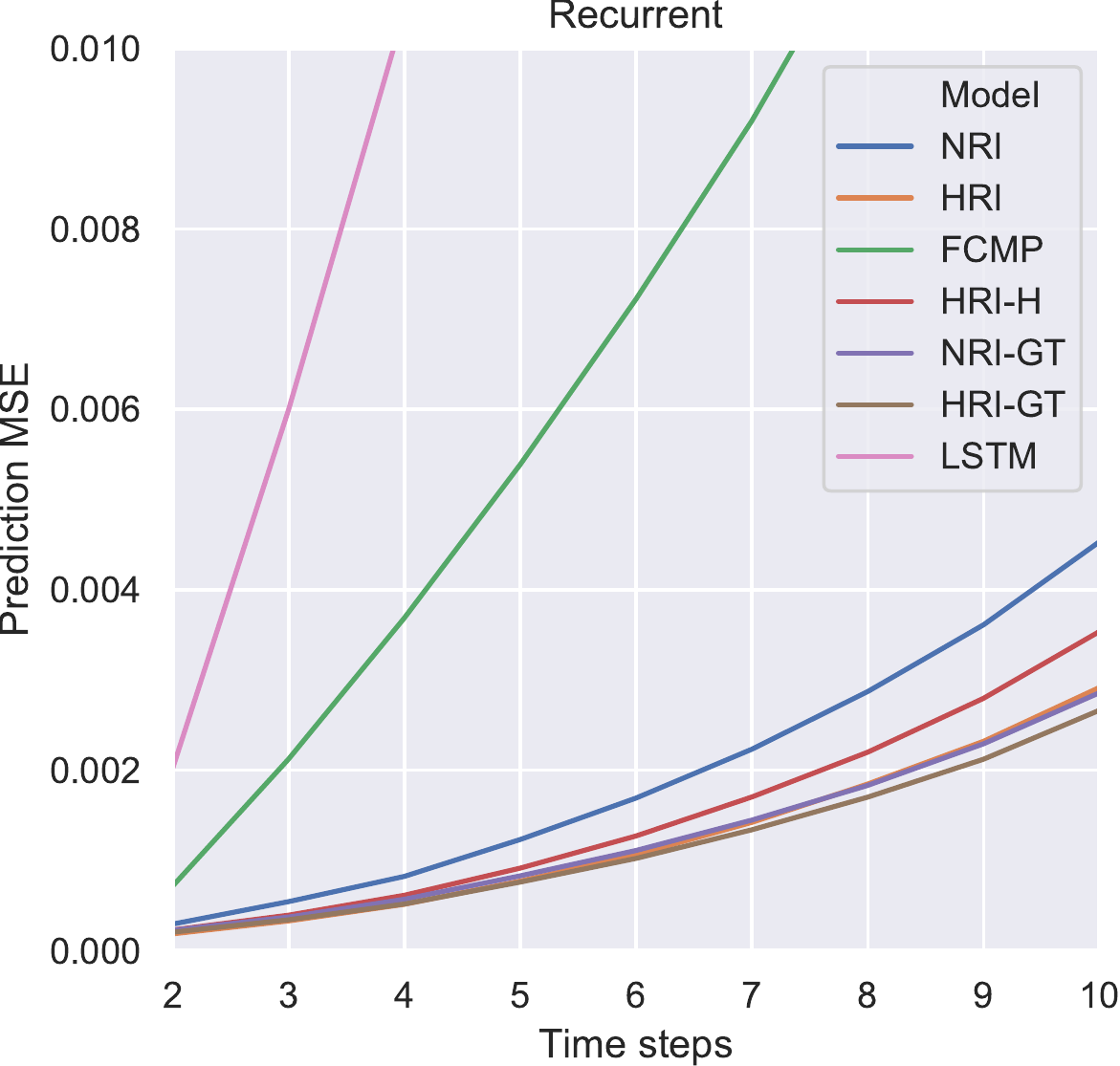}
    \caption{}
    \label{fig:traj-4-3-lstm-mse}
\end{subfigure}
\caption{ 
Performance on 4-3-state-springs. 
We compare HRI to (a) baselines and (b) ablations in terms of the ``normalized'' negative log likelihood (higher is better). 
(c) MSE for future prediction (prediction rollouts).
} 
\label{fig:traj-4-3}
\end{figure*}

\section{Related Work}

More generic approaches to future frame prediction are typically based on RNNs, which are either optimized for next-step prediction directly~\cite{srivastava2015unsupervised,lotter2016deep}, e.g. using a variational approach~\cite{babaeizadeh2017stochastic,denton2018stochastic, kumar2019videoflow}, or (additionally) require adversarial training~\cite{lee2018stochastic,vondrick2016generating}.
Several such approaches were also proposed in the context of physical reasoning~\cite{finn2016unsupervised, lerer2016learning, li2016fall}.
However, unlike our approach, they do not explicitly distinguish between objects and relations.
This is known to affect their ability to accurately model physical interactions and extrapolation~\cite{van2018relational,garnelo2019reconciling}, despite their remarkable capacity for modeling more complex visually scenes.\looseness=-1

More closely related approaches to physical prediction explicitly distinguish object representations and incorporate a corresponding relational inductive bias, typically in the form of a graph network~\cite{battaglia2018relational}.
In this case, reasoning is based on message-passing between object states and relations are either inferred heuristically~\cite{chang2016compositional,mrowca2018flexible,lingelbach2020towards}, implicitly by the message passing functions~\cite{sukhbaatar2015end, battaglia2016interaction, santoro2017simple, watters2017visual, sanchez2018graph, sanchez2020learning} e.g. via attention~\cite{hoshen2017vain, van2018relational}, or explicitly as in NRI~\cite{kipf2018neural}.
Typically, these approaches assume access to supervised state descriptions, and only few works also infer object representations from raw visual images~\cite{van2018relational,veerapaneni2019entity,watters2019cobra}.
RIMs~\cite{goyal2019recurrent} impose a modular structure into an RNN, whose sparse communication can be seen as a kind of relational inference.
However, neither of these additionally incorporate a hierarchical inductive bias to cope with more complex physical interactions (and corresponding objects).
Arguably, this is due to the difficulty of inferring corresponding part-based object representations, and indeed prior approaches that do incorporate a hierarchical inductive bias rely on supervised state descriptions~\cite{mrowca2018flexible,lingelbach2020towards}.

Related approaches that specifically focus on the issue of learning object representations from raw visual observations typically make use of mixture models~\cite{greff2017neural,greff2019multi} or attention~\cite{eslami2016attend,kosiorek2018sequential,stanic2019rsqair,burgess2019monet,crawford2019spatially, jiang2019scalable}.
In our case, we adopt an approach based on spatial slots~\cite{van2019perspective,greff2020binding} that puts less emphasis on the exact correspondence to objects.
Unlike prior work~\cite{santoro2017simple,zambaldi2018deep}, we demonstrate how spatial slots can be extended to a hierarchical setting.%

Hierarchical structure learning has previously been explored in vision.
In this case, hierarchies are either learned from 3D volumetric primitives~\cite{tulsiani2017learning}, or using ground truth pixel-wise flow fields to guide object or object parts segmentation~\cite{xu2019unsupervised}.
In concurrent work~\citep{mittal2020learning}, RIMs were extended to consider a hierarchy of modules, although their correspondence to perceptual objects (and parts) is unclear.
A more relevant approach was recently proposed by \cite{deng2019generative}, which is able to infer a hierarchy of objects and parts, directly from raw visual images.
However, it was not explored how this approach can be adapted to video, and how these abstractions can be used for physics prediction.

\begin{figure*}[t]
\centering
\begin{subfigure}[b]{0.29\textwidth}  
    \centering 
    \includegraphics[width=\textwidth]{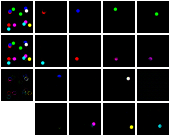}
    \caption{}
    \label{fig:imgs-nll-4-3-slots}
\end{subfigure}
\hspace{0.01\textwidth}
\begin{subfigure}[b]{0.39\textwidth}  
    \centering 
    \includegraphics[width=\textwidth]{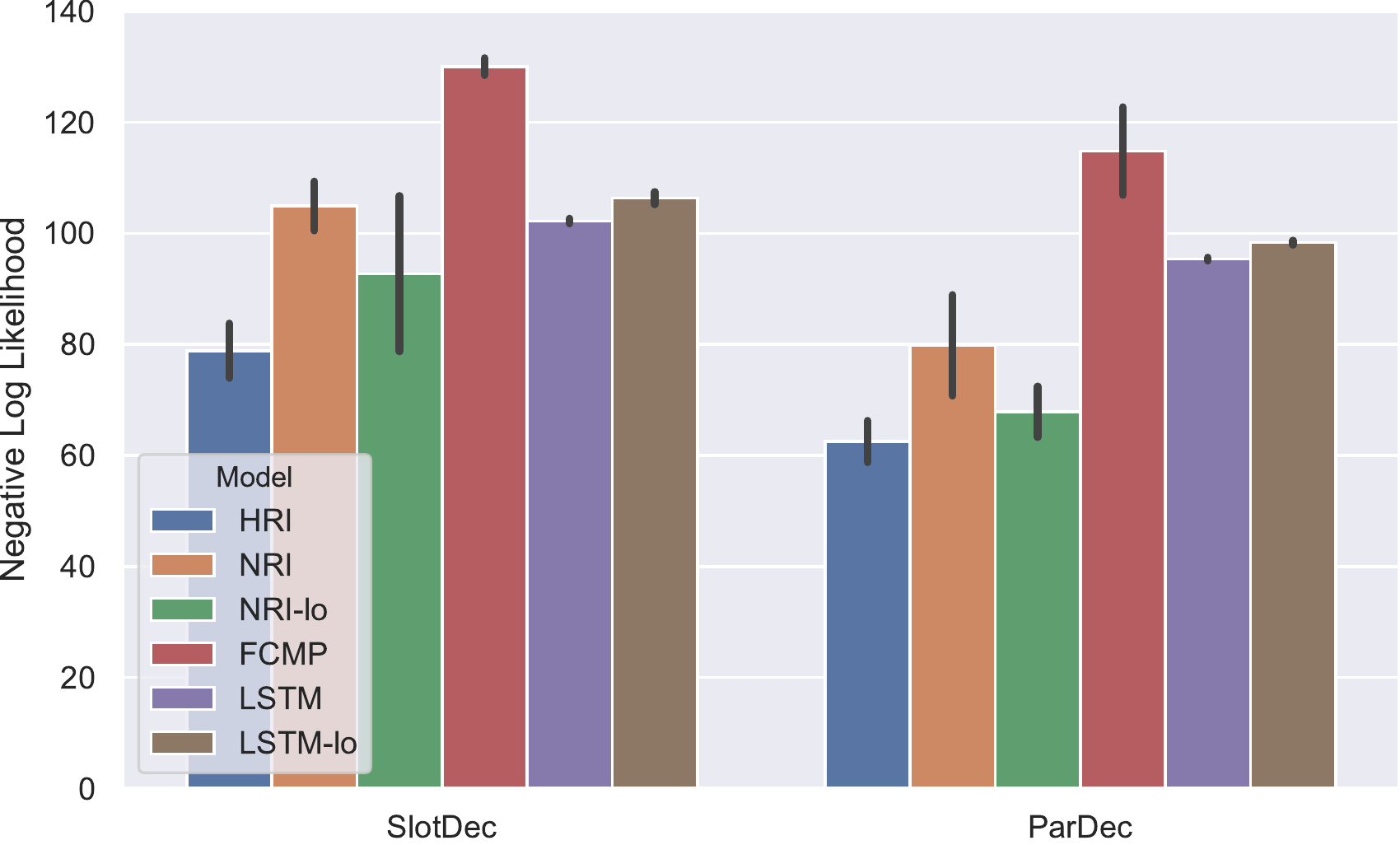}
    \caption{}
    \label{fig:imgs-nll-4-3-springs}
\end{subfigure}
\hspace{0.01\textwidth}
\begin{subfigure}[b]{0.17\textwidth}
    \centering
    \includegraphics[width=\textwidth]{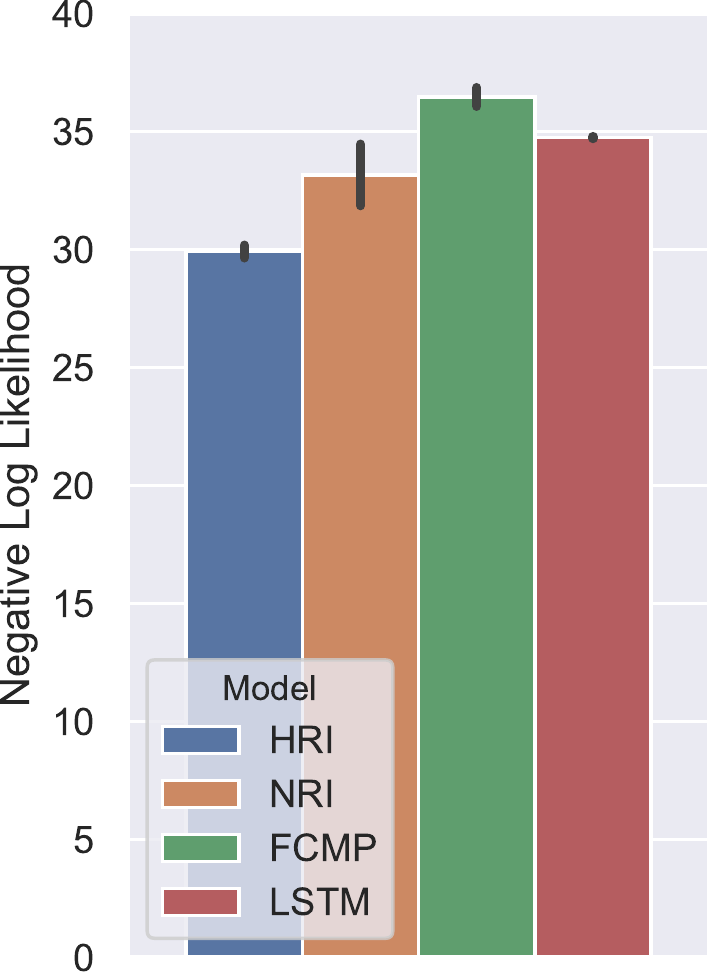}
    \caption{}
    \label{fig:imgs-nll-4-3-hmd}
\end{subfigure}
\caption{ 
(a) HRI introspection: first column contains ground-truth, prediction and their difference. The other columns show the 16 object slots decoded separately. 
(b) Negative log likelihood for all models on the 4-3-visual-springs and (c) Human3.6M.} 
\label{fig:imgs-nll-4-3}
\end{figure*}

\section{Experiments}

In this section we evaluate HRI on four different dynamics modelling tasks: state trajectories of objects connected via finite-length springs in a hierarchical structure (\emph{state-springs}); corresponding rendered videos (\emph{visual-springs}); rendered joints of human moving bodies (\emph{Human3.6M}); and raw videos of moving humans (\emph{KTH}).
We compare HRI to NRI~\cite{kipf2018neural}, which performs relational inference but lacks a hierarchical inductive bias, and to an LSTM~\cite{hochreiter1997long} that concatenates representations from all objects and predicts them jointly, but lacks a relational inference mechanism altogether.
\cref{sec:appendix:experiment-details} contains all experimental details.
Reported results are mean and standard deviations over 5 seeds.

\subsection{State Springs Dataset}

\begin{figure}[h]
    \centering
    \includegraphics[width=0.8\columnwidth]{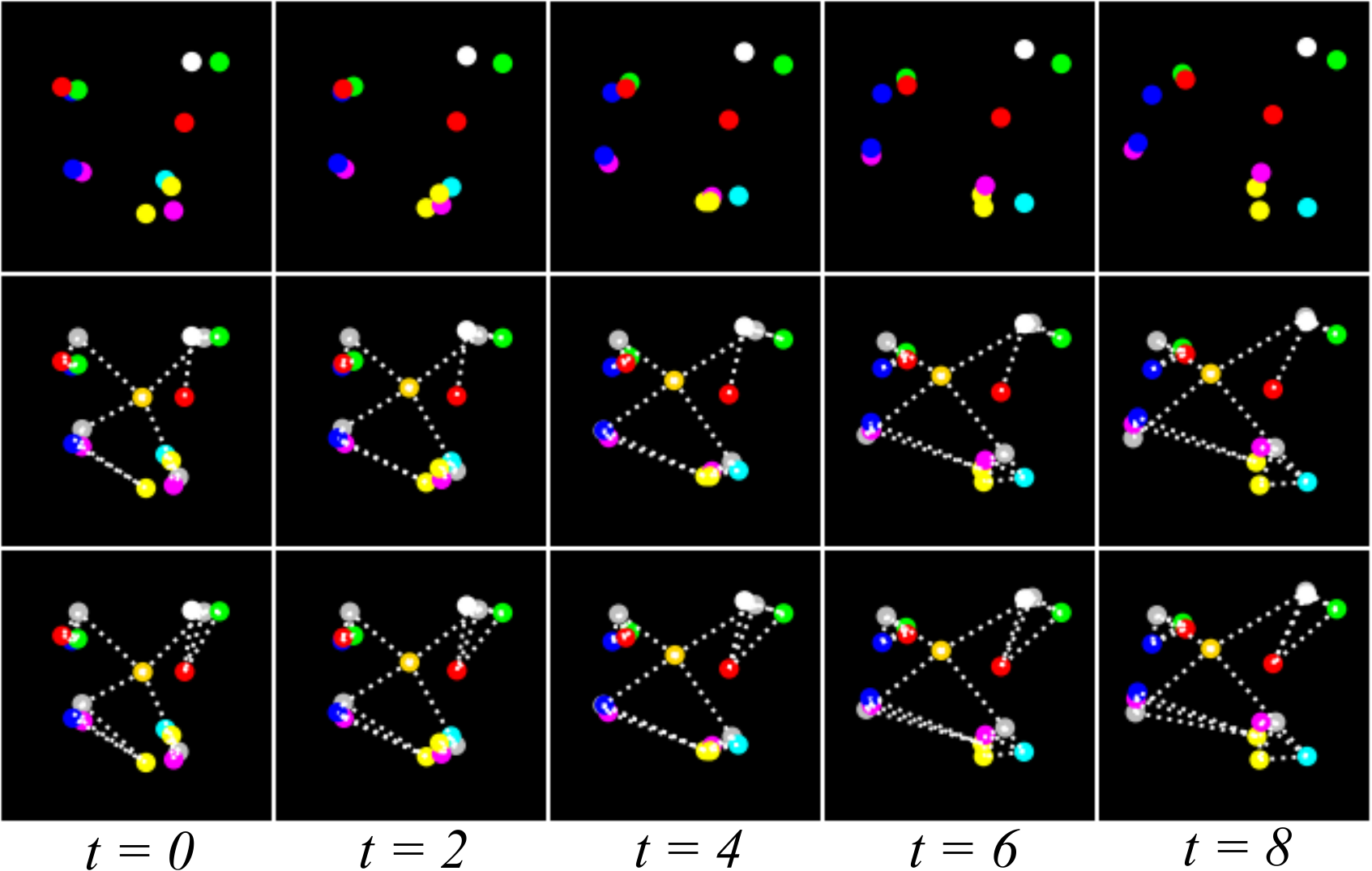}
    \caption{ 
    Rendered inputs for \emph{4-3-state-springs} (leaf objects only) (top); 
    full interaction graph, unobserved by model (middle); 
    Predictions and inferred edges by HRI (bottom).}
    \label{fig:rendered-traj}
\end{figure}

We consider synthetic physical systems containing simple objects connected via finite-length springs that can be organized according to a hierarchical interaction graph (Figure~\ref{fig:rendered-traj}, middle row).
Here, an approach that attempts to model the underlying system dynamics (which are highly complex) would clearly benefit from a hierarchical inductive bias, which allows us to validate our design choices.
In all configurations we consider hierarchies with 3 levels (containing a root node, intermediate nodes, and leaf nodes), whose connectivity is decided randomly while ensuring that all nodes are connected.
Corresponding physical objects are initialized at random locations with the root node biased towards the center and the intermediate and leaf nodes towards a particular quadrant to reduce clutter (see also Appendix~\ref{sec:appendix:experiment-details}). 

We experiment with hierarchies containing 4 intermediate nodes, each having 3 or 4 leaf nodes, denoted as \emph{4-3-state-springs} and \emph{3-3-state-springs}, respectively.
Each model receives as input the 4-dimensional state trajectories: $x(t), y(t), \Delta x(t), \Delta y(t)$.

\paragraph{Comparison to baselines}

We compare HRI to NRI and LSTM on \emph{4-3-state-springs} (Figure~\ref{fig:traj-4-3-baselines-norm-nll}) and \emph{3-3-state-springs} (Figure~\ref{fig:traj-3-3-baselines-norm-nll} in \cref{sec:appendix:additional-results}), in terms of the negative log likelihood inversely proportional to a version of HRI that operates on the ground-truth interaction graph (HRI-GT)\footnote{This allows us to factor out the complexity of the task and make it easier to compare results \emph{between} tasks.}.
In this case, values closer to 1.0 are better, although we also provide raw negative log likelihoods in \cref{fig:traj-4-3-nll}.\looseness=-1 %
It can be observed that HRI markedly outperforms NRI on this task, and that both significantly improve over the LSTM (which was expected). 
These findings indicate that the hierarchical inductive bias in HRI is indeed highly beneficial for this task.

We also compare to baselines that focus only on the leaf objects (\emph{NRI-lo} and \emph{LSTM-lo}) and are therefore not encouraged to separate hierarchical sources of influence (i.e. we impose less structure).
It can be observed that this does not result in better predictions and the gap between NRI-lo and NRI indicates that explicitly considering multiple levels of abstraction, as in HRI, is indeed desirable.

\begin{figure*}[t]
\centering
\begin{subfigure}[b]{0.55\textwidth}  
    \includegraphics[width=\textwidth]{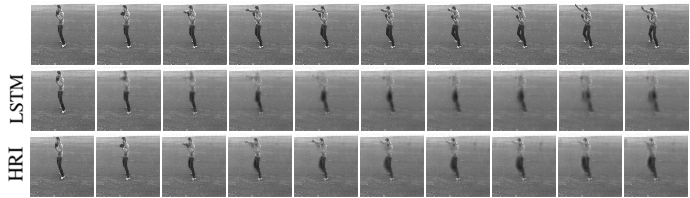}
    \caption{}
    \label{fig:kth-rollout}
\end{subfigure}
\hspace{0.02\textwidth}
\begin{subfigure}[b]{0.11\textwidth}  
    \centering 
    \includegraphics[width=\textwidth]{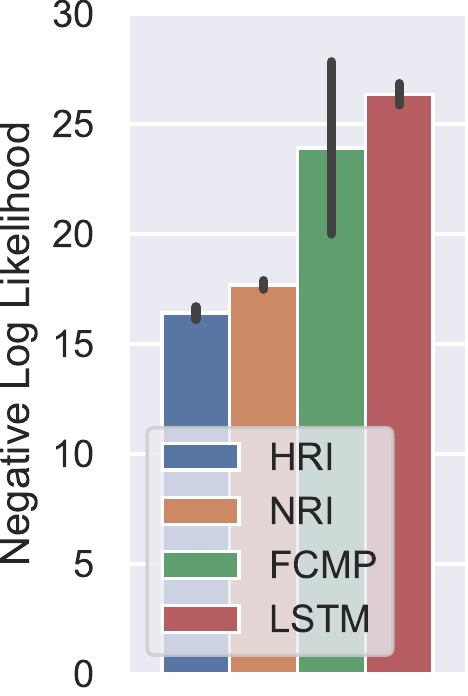}
    \caption{}
    \label{fig:kth-bar}
\end{subfigure}
\hspace{0.02\textwidth}
\begin{subfigure}[b]{0.18\textwidth}
    \centering
    \includegraphics[width=\textwidth]{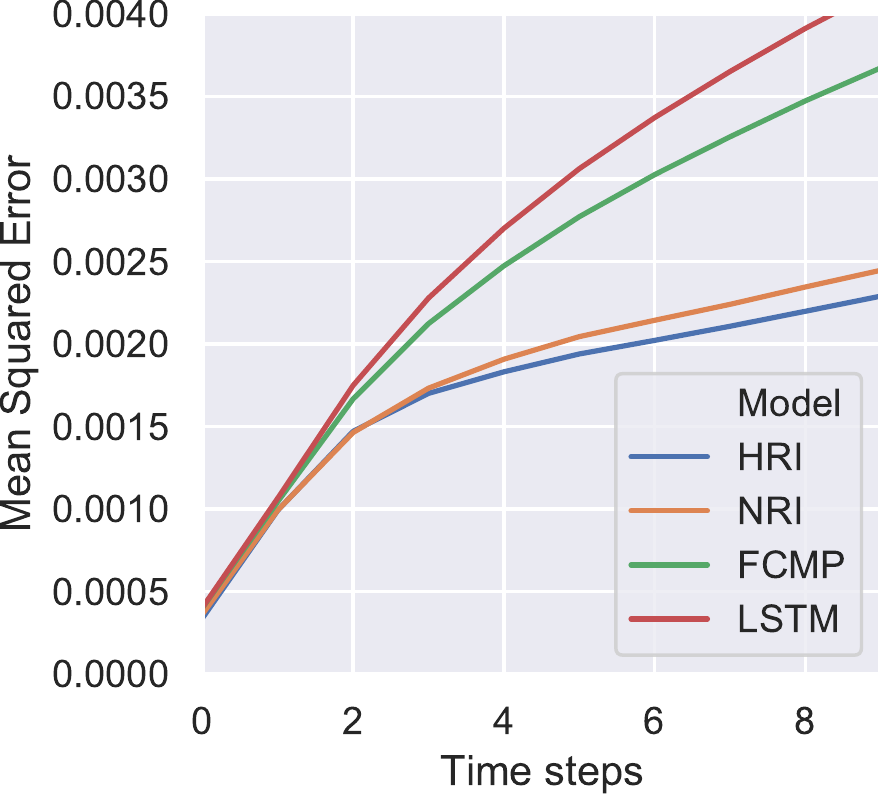}
    \caption{}
    \label{fig:kth-line}
\end{subfigure}
\caption{ 
(a) Ground truth (top) and predicted 10 time steps rollout for LSTM (middle) and HRI (bottom) row on KTH. 
(b) Negative log likelihood for all models on KTH. 
(c) Rollout predictions accuracy for all models on KTH.} 
\label{fig:kth}
\end{figure*}

\paragraph{Ablation \& Analysis}

We conduct an ablation of HRI, where we modify the relational inference module.
We consider FCMP, a variation of NRI that does not perform relational inference but assumes a fixed fully connected graph, and HRI-H, which is given knowledge about the `valid' edges in the ground-truth hierarchical graph to be inferred and performs relational inference only on those.

In Figure~\ref{fig:traj-4-3-ablation-norm-nll} and Figure~\ref{fig:traj-3-3-ablation-norm-nll} (\cref{sec:appendix:additional-results}) it can be observed how the lack of a relational inference module further harms performance.
Indeed, in this case it is more complex for FCMP to implement a similar solution, since it requires ``ignoring'' edges between objects that do not influence each other.
It can also be observed how HRI outperforms HRI-H (but see Figure~\ref{fig:traj-3-3-ablation-norm-nll} where they perform the same),  which is surprising since the latter essentially solves a simpler task.
We speculate that training interactions could be responsible for this gap.
Switching to static graph inference yielded a significant drop in performance for NRI ($\sim74$\% worse), while HRI remained mostly unaffected ($\sim5$\% worse).

We also consider the benefit of hierarchical message-passing in isolation by comparing HRI to NRI-GT, which receives the ground-truth interaction graph.
In Figure~\ref{fig:traj-4-3-ablation-norm-nll} and Figure~\ref{fig:traj-3-3-ablation-norm-nll} (\cref{sec:appendix:additional-results}) it can be seen how the lack of hierarchical message-passing explains part of the previously observed gap between HRI and NRI, but not all of it.
It suggests that by explicitly considering multiple levels of abstraction (as in HRI), conducting relational inference generally becomes easier on these tasks.
Finally, we consider the effect of using a feedforward dynamics predictor (\cref{fig:traj-4-3-nll,fig:traj-3-3-nll} in Appendix~\ref{sec:appendix:additional-results}).  
It can be seen that in the feedforward case, position and velocity are not enough to derive interactions between objects.
For this reason we exclusively focus on the recurrent predictor from now on.

\paragraph{Long-term Prediction and Graph Inference}

We investigate if HRI is able to perform long-term physical predictions by increasing the number of rollout steps at test-time. 
In Figure~\ref{fig:traj-4-3-lstm-mse} we report the MSE between the predicted and ground truth trajectories and compare to previous baselines and variations.
It can be observed that HRI outperforms all other models, sometimes even HRI-GT, and this gap increases as we predict deeper into the future.
To qualitatively evaluate the plausibility of HRI's rollout predictions, we generated videos by rendering them. 
As can be seen in Figure~\ref{fig:rendered-traj} (bottom row) both the predicted trajectories and the inferred graph connectivity closely matches the ground-truth.
Similar results are obtained for \emph{3-3-state-springs} in Appendix~\ref{sec:appendix:additional-results}.

\subsection{Visual Datasets}

To generate visual data for springs we rendered \emph{4-3-state-springs} and \emph{3-3-state-springs} with all balls having the same radius and one of the 7 default colors, assigned at random\footnote{
Similar results were obtained when using homogeneous colors (\cref{fig:imgs-nll-bw-3-3,fig:imgs-nll-bw-4-3}) and when varying shape (\cref{fig:imgs-nll-4-3-v23,fig:imgs-nll-4-3-v24,fig:imgs-nll-4-3-v29,fig:imgs-nll-4-3-v20-08,fig:imgs-nll-4-3-v20c}).
}.
Additionally, we consider \emph{Human3.6M}~\cite{ionescu2013human3} (using rendered joints) and \emph{KTH}~\cite{schuldt2004recognizing} (using raw videos).
We train each model in two stages, which acts as a curriculum: first we train the visual encoder and decoder on a reconstruction task, afterwards we optimize the dynamics parameters on the prediction task.

\paragraph{Visual Springs}

To create the visual springs dataset, we render only the \emph{leaf nodes}.
Hence the visible balls themselves are the ``parts'' that must be grouped in line with more abstract ``objects'' corresponding to the intermediate and root nodes of the interaction graph (e.g. the gold and silver balls in the bottom row of \cref{fig:rendered-traj}) which are not observed.

In Figure~\ref{fig:imgs-nll-4-3-springs} we compare HRI to several baselines and previously explored variations  (the results for \emph{3-3-visual-springs} are available in Figure~\ref{fig:imgs-nll-3-3} in Appendix~\ref{sec:appendix:additional-results}).
We are able to observe similar results, in that HRI is the best performing model, although the added complexity of inferring object states results in smaller differences.
This is better understood by comparing the difference between HRI and NRI\footnote{Note that NRI requires object states. Here, we consider an adaptation using the encoder and decoder proposed for HRI.} to the difference between HRI and NRI-lo. 
It can be seen that NRI-lo performs slightly worse than HRI (although it is less stable) and is much better than NRI, which suggests that inferring the intermediate and root node objects is one of the main challenges in this setting.
Notice, however, that HRI continues to markedly improve over the LSTM and visual NRI in this setting.
We note that comparing to NRI-GT or HRI-GT is not possible, since the mapping from learned object representations to those in the graph is unknown.

Comparing SlotDec and ParDec in Figure~\ref{fig:imgs-nll-4-3-springs}, we observe that better results are obtained using the latter.
We suspect that this is due to the visual occlusions that occur, which makes summation in pixel space more difficult.
Nonetheless, in most cases we observe a clear correspondence between spatial slots and information about individual objects (see Figure~\ref{fig:imgs-nll-4-3-slots}).
Finally, Figure~\ref{fig:rollout} (Appendix \ref{sec:appendix:additional-results}) demonstrates how future predictions by HRI match the ground-truth quite well.
Similarly, we observe that the inferred interaction graph by HRI resembles the ground-truth much more closely compared to NRI (\cref{fig:springs-inferred-graphs} in \cref{sec:appendix:additional-results}).

When evaluating HRI trained on \emph{4-3-visual-springs} on \emph{3-3-visual-springs} (i.e. when \emph{extrapolating}) the relative ordering is preserved (\cref{fig:imgs-nll-4-3-gen-to-3-3} in \cref{sec:appendix:additional-results}).

\paragraph{Real-world Data}

We consider \emph{Human3.6M}, which consists of 3.6 million 3D human poses composed of 32 joints and corresponding images taken from professional actors in various scenarios. 
Here we use the provided 2D pose projections to render 12 joints in total (3 per limb) as input to the model.
This task is significantly more complex, since the underlying system dynamics are expected to vary over time (i.e. they are non-stationary). \cref{fig:imgs-nll-4-3-hmd} demonstrates the performance of HRI and several baselines on this task. 
Note that HRI is the best performing model, although the gap to NRI and LSTM is smaller. 
This can be explained by the fact that many motions, such as when sitting, eating, and waiting involve relatively little motion or only motion in a single joint (and thereby lack hierarchical interactions).
Example future predictions by HRI can be seen in \cref{fig:h36m-rollout}.\looseness=-1

Finally, we apply HRI directly on the \emph{raw} videos of \emph{KTH}, which consists of people performing one of six actions (walking, jogging, running, etc.).
Figure~\ref{fig:kth-rollout} shows a comparison between the LSTM baseline and HRI model.
In the video generated by the LSTM, the shape of the human is not preserved, while HRI is able to clearly depict human limbs.
We prescribe this to the ability of HRI to reason about the objects in the scene in terms of its parts and their interactions, which in turn simplifies the physics prediction.
This observation is also reflected in the quantitative analysis in \cref{fig:kth-bar,fig:kth-line}, where large differences can be observed, especially when predicting deeper into the future.

\section{Conclusion}

Hierarchical Relational Inference (HRI) is a novel approach to common-sense physical reasoning capable of learning to discover objects, parts, and their relations, directly from raw visual images in an unsupervised fashion.
It builds on the idea that the dynamics of complex objects are best modeled as hierarchies of parts that separate different sources of influence.
We compare HRI's predictions to those of a strong baseline on synthetic and real-world physics prediction tasks where it consistently outperforms.
Several ablation studies validate our design choices.
A potential limitation of HRI is that it groups parts into objects based on spatial proximity, which may be suboptimal in case of severe occlusions.
Further, it would be interesting to provide for inter-object interactions, and unknown hierarchy depths.
Both might be handled by increasing the hierarchy depth, and relying on the relational inference module to ignore non-existent edges.

\section*{Acknowledgments}
We thank Anand Gopalakrishnan, R{\'o}bert Csord{\'a}s, Imanol Schlag, Louis Kirsch and Francesco Faccio for helpful comments and fruitful discussions.
This research was supported by the Swiss National Science Foundation grant 407540\_167278 EVAC - Employing Video Analytics for Crisis Management, and the Swiss National Science Foundation grant 200021\_165675/1 (successor project: no: 200021\_192356).
We are also grateful to NVIDIA Corporation for donating several DGX machines to our lab and to IBM for donating a Minsky machine.

{
\small
\bibliography{biblio}
}

\appendix
\clearpage
\appendix

\section{Experiment Details}
\label{sec:appendix:experiment-details}

In this section we decribe all datasets and the training procedure in detail.
Additionally, we provide model architecture configurations for the HRI model, ablation models, NRI and LSTM baselines.
Reported results in the bar plots are the mean and standard deviation obtained for each model using $5$ different random seeds.

\subsection{Datasets}

Four main datasets are used: \emph{state-springs}, \emph{visual-springs} (rendered version of state-springs), and two real-world benchmark datasets: one obtained by rendering the joints of the \emph{Human3.6M} dataset and the other using the raw videos of the \emph{KTH} dataset.

\subsubsection{Springs}

We simulate a system of moving objects connected via finite-length springs starting from the open source simulator implementation of~\citet{kipf2018neural}.
We make the following modifications: objects are connected via finite length springs (instead of ideal springs) and the sampled graphs have a hierarchical structure in terms of connectivity, initial spatial positions and the spring constants (which reflects in the speed by which objects in different layers of the hierarchy move).
Objects are connected via finite springs, which makes them act according to a modified Hooke's law:
\begin{equation}
    F_{ij} = -k_F (r_i - r_j - l \cdot \frac{r_i - r_j}{ | r_i - r_j | }),
\end{equation}
where $r_i$ and $r_j$ are $(x,y)$-coordinates of objects $i$ and $j$, $k_F$ is the spring constant and $l$ is its length.
The objects are connected in a hierarchical graph, and they move inside a unit box (bounce elastically from the walls).
To simulate the system, we initialize the root node position randomly in a neighborhood around the center of the image by sampling its $(x,y)$ coordinates from $\mathcal{N}(0,0.25)$.
We then initialize the intermediate and the leaf nodes randomly inside each of the four image quadrants to ensure an initial bias towards spatial grouping.
In particular, we sample from Gaussian distributions with variance $0.25$ and the means (for $(x,y)$-coordinates) being the centers of the four quadrants: $(-0.25, 0.25)$, $(0.25, 0.25)$, $(-0.25, -0.25)$ and $(0.25, -0.25)$.
For each sample in the dataset, we sample a graph with \emph{random} connectivity: we start from a full tree graph, where sibling nodes are fully connected, then drop edges at random with a probability of $0.5$, but ensure that the resulting graph is connected.
The spring lengths are: $0.4$ between the root and intermediate nodes, $0.1$ between intermediate and leaf nodes, $0.65$ within intermediate node siblings and $0.2$ within leaf node siblings.
All springs have the same constant, except for springs between leaf node siblings, which have a value that is half the value of other constants.
In total we generate a dataset of $5 \cdot 10^5$ training, $10^5$ validation, and $10^5$ test sequences, each $50$ frames (steps) long.

We note that all objects in the springs dataset are free to move anywhere in the space, and not limited to a particular quadrant. 
The intermediate nodes are only initialized with a bias towards quadrants, which was done to obtain a scenario where the objects naturally group together from the beginning of the sequence. 
This is identical to initializing all positions at random, and then performing rollouts until the objects' positions converge (which would always happen due to the hierarchical forces that act between them).

\subsubsection{Human3.6M}

This dataset~\cite{ionescu2013human3} consists of 3.6 million 3D human poses, corresponding 2D pose projections and images.
We use the provided 2D pose projections to render 12 joints in total (3 of each limb). 
In total there are 11 professional actors in 17 scenarios, from which subjects number 1, 5, 6, 7, and 8 are used for training, and subjects number 9 and 11 for testing.
This allows us to create 10k training and 3.5k test sequences of 50 frames.

\subsubsection{KTH}

The KTH Action dataset~\cite{schuldt2004recognizing} consists of real-world videos of people performing one of six actions: walking, jogging, running, boxing, hand-waving and hand-clapping.
We trained all models on $64 \times 64$ video sequences, by conditioning on 10 frames and training the model to predict the next 10 frames in ten frames in the sequence, as done in previous work~\cite{denton2018stochastic}.
We split the dataset into a training dataset containing 8k sequences and test dataset containing 1.5k test sequences, consisting of subjects not present in the training dataset.

\subsection{Training Details}

All models are trained with Adam
~\cite{kingma2015AdamAM} using default parameters and a learning rate of $5 \cdot 10^{-4}$.
We use a batch size of $32$ and train models for $100$ epochs.
On the visual task we train each model in two stages, which acts as a curriculum: 
for the first $50$ epochs we train the visual encoder and decoder on a reconstruction task, afterwards we optimize the relational module and dynamics parameters on the prediction task.

Optimizing only for the next step prediction task can lead to a predictor which ignores the inferred relational graph (also noted in~\cite{kipf2018neural}).
To avoid this, in the feedforward case we predict $10$ steps in the future by feeding in predicted output as the next step input, and only feed the ground truth input every $10$ steps. 
In the recurrent case this might lead to instabilities, so we introduce a ``burn-in-phase'', where we feed the ground truth input at the beginning, and predict the last $10$ steps of the sequence.
To train the models we optimize for ELBO over such sequences, whereas we evaluate the models on the next step prediction task.
The Gaussian output distribution has a fixed variance $\sigma^2=5 \cdot 10^{-5}$.

 We evaluate the models based on the negative log likelihood loss and the ``normalized'' negative log likelihood (\cref{fig:traj-4-3-baselines-norm-nll,fig:traj-4-3-nll}), which is inversely proportional to a version of HRI that operates on the ground-truth interaction graph (HRI-GT) (higher value is better).
The latter allows us to factor out the complexity of the task and make it easier to compare results \emph{between} tasks.
We average the negative log likelihood either over the number of objects (for the state datasets) or pixels (for the visual datasets).

\subsection{HRI Architecture Details}

We describe every module in terms of blocks it consists of, e.g. node-to-edge MLP, edge-to-node MLP, LSTM etc.
The architectures in all tables represent the forward pass, except for~\cref{app:visual-encoder-arch}, where for the first part (leaf object encoder) we have two variants (depending on whether we use SlotDec or ParDec), whereas the hierarchy inference module is the same in both cases.

A high-level overview of the HRI model is presented in~\cref{fig:hri}, with a high-level summary of all components.
Below we describe each component in detail.

\paragraph{Visual Encoder}

The visual encoder takes as input the concatenation of $32\times 32 \times 3$ RGB frame and a $32\times 32 \times 2$ $(x,y)$-fixed coordinate channels (as in~\citet{liu2018intriguing,watters2019spatial}, simplifying the object's position inference relative to the image frame), processes it with several convolutional layers with ReLU non-linearities and batch norm ~\cite{ioffe2015batch} and outputs a hierarchy of object representations.
First it infers the $4 \times 4$ leaf objects (note different variants for this part in~\cref{app:visual-encoder-arch} depending on the decoder), from which $4$ intermediate nodes and $1$ root node are inferred.
This results in 16 leaf objects, 4 intermediate objects, and one root object, each 48-dimensional.
They are all mapped with a FC layer (with shared weights) to 16-dimensional vectors and fed through another (Object-wise VAE) FC layer ($\mu$ and $\sigma$ for the standard VAE reparametrization).
For KTH experiments the input resolution is $64\times 64 \times 1$, handled by repeating the first convolutional layer of the encoder.

\begin{table}[h]
    \centering
    \caption{\textbf{Visual encoder architectures}}
    \begin{tabular}{l}
        \toprule
        \textbf{Leaf Object Encoder for SlotDec} \\
            \midrule 
            $8\times 8$ conv, $48$ ReLU units, stride $8$, batch norm
            \\
            \midrule
        \textbf{Leaf Object Encoder for ParDec} \\
            \midrule 
            $8\times 8$ conv, $48$ ReLU units, max pool $2$, batch norm \\
            $8\times 8$ conv, $48$ ReLU units, max pool $2$, batch norm \\
            $8\times 8$ conv, $48$ ReLU units, max pool $2$, batch norm \\
            \midrule 
        \textbf{Hierarchical objects inference} \\
            \midrule 
            $2\times 2$ conv, $48$ ReLU units, max pool $2$, batch norm \\
            $2\times 2$ conv, $48$ ReLU units, max pool $2$, batch norm \\
            Object-wise FC, $16$ ReLU units. \\
            Object-wise VAE FC, $2 \times 16$ units ($\mu, \sigma $). \\
        \bottomrule
    \end{tabular}
    \label{app:visual-encoder-arch}
\end{table}

\paragraph{Relational Inference Module}

The relational inference module takes a sequence of $K=10$ object vectors (16-dimensional in visual and 4-dimensional in state case) as input and outputs one-hot encoding (2-dimensional) of the pair-wise relations $\be_{(i,j)}$ (the outputs of the $f_o$ MLP in\cref{app:relational-inference-arch}).
The samples are drawn as $\br_{ij} = \mathrm{softmax}((\be_{(i,j)} + \mathbf{g})/\tau)$ where $\mathbf{g}$ is drawn from a $\mathrm{Gumbel}(0,1)$ distribution and $\tau=0.5$ is the temperature parameter.

\begin{table}[h]
    \centering
    \caption{\textbf{Relational inference module architectures}}
    \begin{tabular}{l}
        \toprule
        \textbf{Node-embeding MLP} \\
            \midrule 
            Concatenate $K$ object states in a slot-wise manner \\
            FC, $64$ ELU \\
            FC, $64$ ELU, batch norm \\
            \midrule
            Concatenate object pairs slot-wise $o_{ij}=[o_i,o_j]$ \\
            \midrule
        \textbf{Node-to-edge MLP $f_{n2e}$} \\
            \midrule 
            FC, $64$ ELU \\
            FC, $64$ ELU, batch norm \\
            \midrule
        \textbf{Edge-to-node MLP $f_{e2n}$} \\
            \midrule 
            FC, $64$ ELU \\
            FC, $64$ ELU, batch norm \\
            \midrule
            Append slot-wise the skip connection of $o_{ij}$ \\
            \midrule
        \textbf{Node-to-edge MLP (shared weights with $f_{n2e}$)} \\
            \midrule 
            FC, $64$ ELU \\
            FC, $64$ ELU, batch norm \\
            \midrule
        \textbf{Output MLP $f_{o}$} \\
            \midrule 
            FC, $64$ ELU \\
            FC, $64$ ELU, batch norm \\
            FC, 2 output units \\
        \bottomrule
    \end{tabular}
    \label{app:relational-inference-arch}
\end{table}

\paragraph{Dynamics Predictor}

The dynamics predictor (~\cref{app:dynamics-predictor-arch}) takes as input inferred object states of dimensionality $d$ and the inferred pairwise edges, and predicts the object states at the next time step via rounds of message passing.
Hierarchical message passing functions $f^{bu}_{MP}, f^{ws}_{MP},$ and $f^{td}_{MP}$ perform a single node-to-edge and edge-to-node message passing operation as in \eqref{eq:gnn_mp}, where their node-to-edge and edge-to-node MLPs all share the same set of weights.
The feedforward variant replaces LSTM with one layer (FC with 64 ReLU units) that takes as input the concatenation of object at the current step and the effect computed by the message passing.

\begin{table}[h]
    \centering
    \caption{\textbf{Dynamics predictor architecture}}
    \begin{tabular}{l}
        \toprule
        \textbf{Bottom-up message passing round $f^{bu}_{MP}$} \\ 
        (on child-parent edges)\\
            \midrule 
        \textbf{Node-to-edge MLP $f_{n2e}$} \\
            FC, $64$ ReLU \\
            FC, $64$ ReLU, batch norm \\
        \textbf{Edge-to-node MLP $f_{e2n}$} \\
            FC, $64$ ReLU \\
            FC, $64$ ReLU, batch norm \\
            \midrule
        \textbf{Within-siblings message passing round $f^{ws}_{MP}$} \\ 
            (on sibling edges) \\
            {Shared weights of $f_{n2e}$ and $f_{e2n}$ MLPs with $f^{bu}_{MP}$} \\
            \midrule 
        \textbf{Top-down message passing round $f^{td}_{MP}$} \\
            (on parent-child edges) \\
            {Shared weights of $f_{n2e}$ and $f_{e2n}$ MLPs with $f^{bu}_{MP}$} \\
            \midrule 
        \textbf{LSTM} \\
            \midrule 
            LSTM, 64 hidden units \\
            \midrule
        \textbf{Output MLP $f_{o}$} \\
            \midrule 
            FC, $64$ ReLU \\
            FC, $d$ output units. \\
        \bottomrule
    \end{tabular}
    \label{app:dynamics-predictor-arch}
\end{table}

\paragraph{Visual Decoder}

The visual decoder takes as input a set of $N$ vectors ($d=16$-dimensonal object states) and produces the output image according to the architecture in \cref{app:visual-decoder-arch}.
For the SlotDec a unique float index $i\in [0,1]$ is appended to each object state, which helps learning the visual object colors, as they are decoded separately as a (permutation invariant) set and then summed.
Note that for the KTH experiments the image resolution is $64\times 64 \times 1$, which is handled by simply adding another `$4\times 4$ convTranspose, 64 ReLU, stride 2' layer to the decoder.
In a similar way we add a convolutional layer to the encoder and by doing this we are able to infer the same hierarchy regardless of the input resolution.
To account for additional visual complexity of predicting 
As in \cite{denton2018stochastic}, we add skip connections between encoder and decoder to enable the model to easily generate static background features, and allow the dynamics predictor to focus on modelling the changes.

\begin{table}[h]
    \centering
    \caption{\textbf{Visual decoder architectures}}
    \begin{tabular}{l}
        \toprule
        \textbf{SlotDec} \\
            \midrule 
            FC, $4 \times d$ ReLU \\
            $4\times 4$ convTranspose, 64 ReLU, stride 2 \\
            $4\times 4$ convTranspose, 64 ReLU, stride 2 \\
            $4\times 4$ convTranspose, 64 ReLU, stride 2 \\
            $4\times 4$ convTranspose, 3 ReLU, stride 2 \\
        \bottomrule
    \end{tabular}
    \begin{tabular}{l}
        \toprule
        \textbf{ParDec} \\
            \midrule 
            $4\times 4$ convTranspose, 64 ReLU, stride 2 \\
            $4\times 4$ convTranspose, 64 ReLU, stride 2 \\
            $4\times 4$ convTranspose, 3 ReLU, stride 2 \\
        \bottomrule
    \end{tabular}
    \label{app:visual-decoder-arch}
\end{table}

\begin{table}[h]
    \centering
    \caption{\textbf{NRI dynamics predictor}}
    \begin{tabular}{l}
        \toprule
        \textbf{Node-to-edge MLP $f_{n2e}$} \\
            \midrule 
            FC, $64$ ReLU \\
            FC, $64$ ReLU, batch norm \\
            \midrule
        \textbf{Edge-to-node MLP $f_{e2n}$} \\
            \midrule 
            FC, $64$ ReLU \\
            FC, $64$ ReLU, batch norm \\
            \midrule
        \textbf{LSTM} \\
            \midrule 
            LSTM, 64 hidden units \\
            \midrule
        \textbf{Output MLP $f_{o}$} \\
            \midrule 
            FC, $64$ ReLU \\
            FC, $d$ output units. \\
        \bottomrule
    \end{tabular}
    \label{app:nri-arch}
\end{table}

\subsection{Ablations}

Following are the ablation-specific configurations:
\begin{itemize}
    \item \textbf{HRI-GT}: HRI model that gets ground truth graph as the input to the dynamics predictor (no relational inference).
    \item \textbf{HRI-H}: HRI model that performs relational inference on a smaller subset of edges (other edges are excluded), by considering the convolutional and pooling operations that infer the hierarchical object slots.
    Let $o1$ be the root object, $o2,o3,o4,o5$ intermediate objects, and $o6-o21$ leaf objects. 
    The subset of edges HRI-H considers are parent-child (and vice-versa child-parent) $(1-2,1-3,1-4,1-5)$, $(2-6,2-7,2-8,2-9)$, $(3-10,3-11,3-12,3-13)$, $(4-14,4-15,4-16,4-17)$ and $(5-18,5-19,5-20,5-21)$ and all within-sibling edges.
    \item \textbf{NRI-GT}: NRI model that gets ground truth graph as the input to the dynamics predictor (no relational inference).
    \item \textbf{FCMP}: NRI model that performs message passing in the dynamics predictor on a fully connected graph (no relational inference).
\end{itemize}

Note also that the performance gain observed in our experiments is not a consequence of an increased model capacity: in the case of visual springs dataset HRI and the NRI baseline have the same number of parameters - 292'141, whereas the LSTM baseline has more - 317'707.

\subsection{NRI Baseline}

To infer the object states on which NRI performs relational inference we use the visual encoder and decoder of the HRI architecture. 
This ensures a fair comparison between NRI and HRI in the visual setting.
We emphasise that standard NRI as presented in \cite{kipf2018neural} did not support learning from visual images.

The dynamics predictor (`decoder' in NRI~\cite{kipf2018neural}) is presented in~\cref{app:nri-arch}, which uses an LSTM~\cite{hochreiter1997long} instead of the GRU~\cite{cho2014learning} cell.

\subsection{LSTM Baseline}

Similarly to the NRI baseline, the LSTM baseline uses the same (pretrained) visual encoder and decoder to map from image to object states, and vice-versa.
We use an LSTM with 64 hidden units that concatenates representations from all objects and predicts their future state jointly.
Essentially, the NRI baseline dynamics predictor can be viewed as extending the LSTM by adding the message passing part (functions $f_{n2e}$ and $f_{e2n}$) based on the inferred interaction graph.
In contrast, the LSTM baseline only explicitly considers the nodes of the graph, but not its edges (relations).

\begin{figure*}[ht]
\centering
\begin{subfigure}[b]{0.48\columnwidth}
    \centering
    \includegraphics[width=\textwidth]{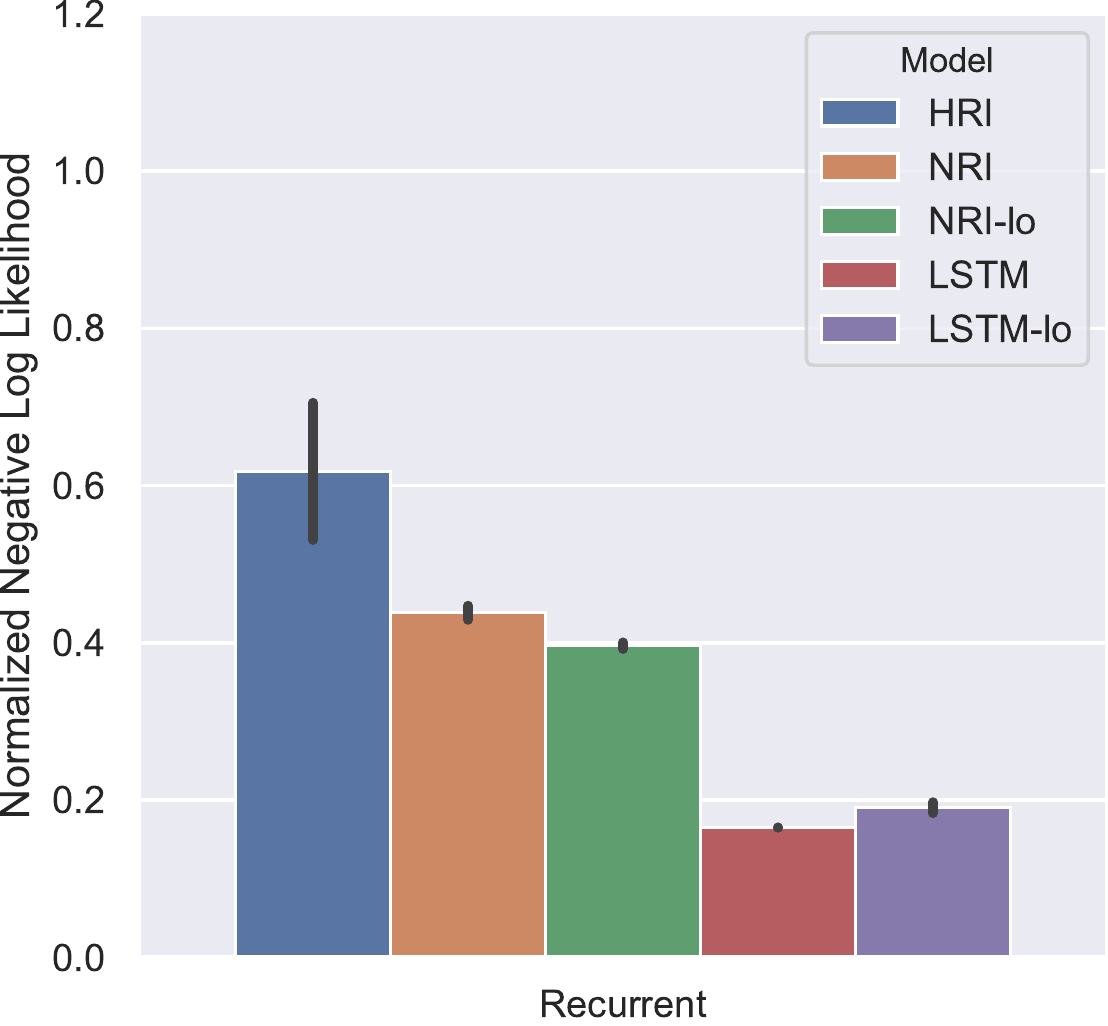}
    \caption{}
    \label{fig:traj-3-3-baselines-norm-nll}
\end{subfigure}
\begin{subfigure}[b]{0.48\columnwidth}   
    \centering 
    \includegraphics[width=\textwidth]{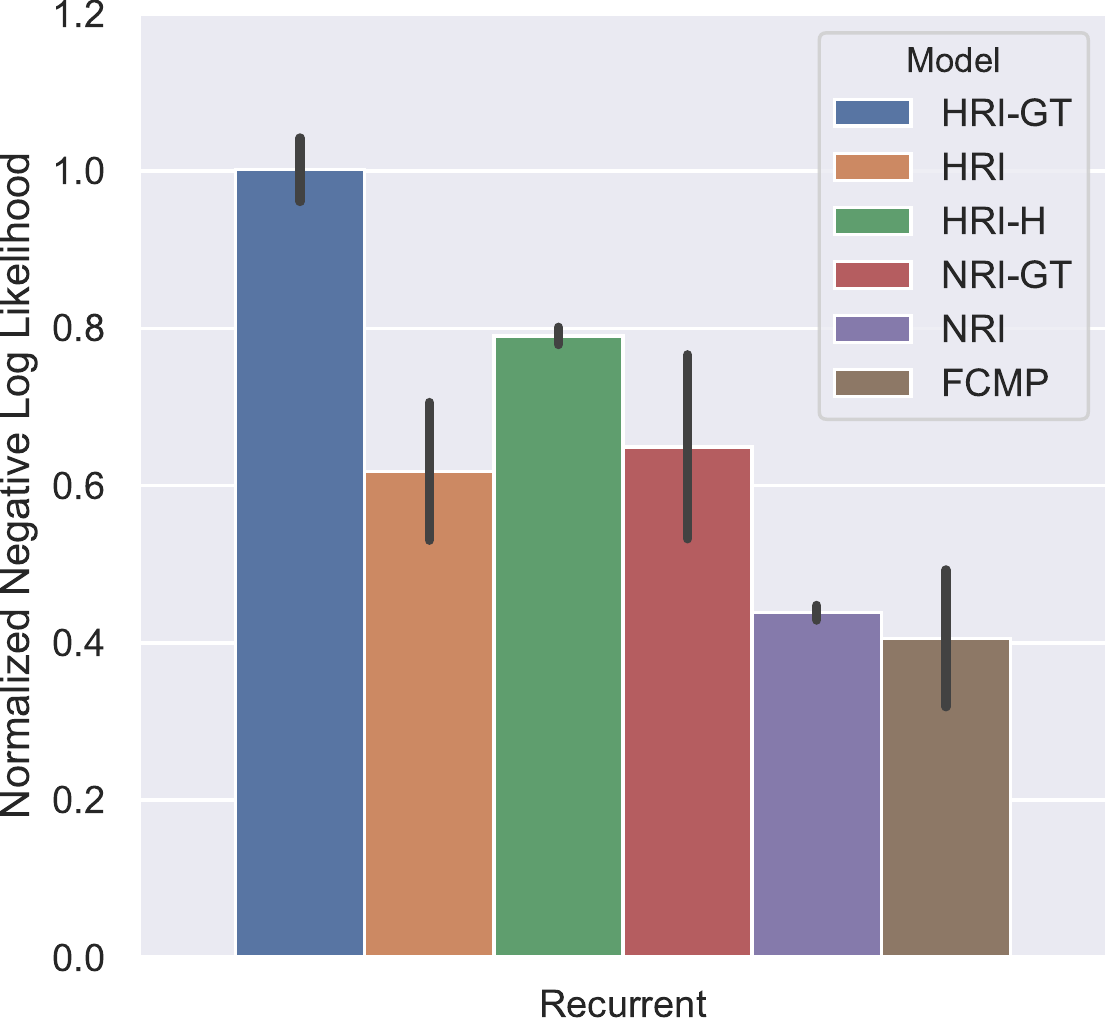}
    \caption{}
    \label{fig:traj-3-3-ablation-norm-nll}
\end{subfigure}
\begin{subfigure}[b]{0.3\textwidth}   
    \centering 
    \includegraphics[width=\textwidth]{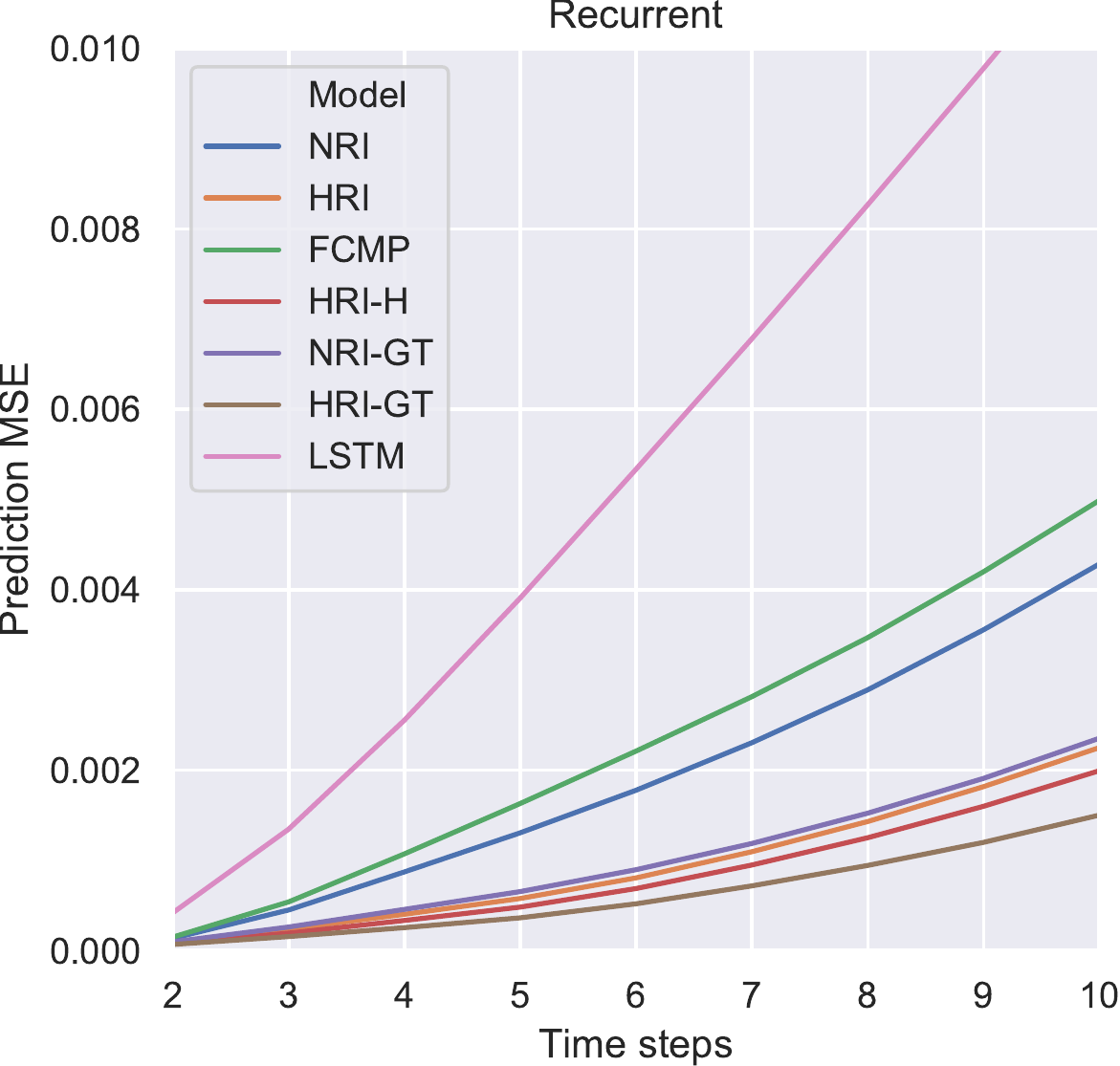}
    \caption{}
    \label{fig:traj-3-3-lstm-mseFULL}
\end{subfigure}
\caption{
Performance on the 3-3-state-springs dataset. 
We compare HRI to (a) baselines and (b) ablations in terms of the ``normalized'' negative log likelihood which is inversely proportional to HRI-GT, which receives the ground-truth graph. 
In this case higher is better. 
(c) MSE when predicting into the future (prediction rollouts).
} 
\label{fig:traj-3-3}
\end{figure*}

\section{Additional Results}
\label{sec:appendix:additional-results}

In this section we provide additional results for the 3-3 state and visual dataset, the results for datasets containing homogeneous (white) balls instead of colored ones: 3-3-state-visual-white and 4-3-state-visual-white in \cref{sec:visual-springs-numbers-bw}, present the rollouts on visual spring datasets and Huma3.6M dataset in~\cref{sec:visual-springs-rollouts-graphs}, results on datasets containing objects of different shapes, sizes and occlusions  in \cref{sec:visual-springs-diverse}, and finally results on generalization (extrapolation) of models trained on 4-3-visual-springs to 3-3-visual springs in \cref{sec:visual-springs-generalizaton}.

\subsection{State Springs}
\begin{figure}[h]
\centering
\begin{subfigure}[b]{0.48\columnwidth}   
    \centering 
    \includegraphics[width=\textwidth]{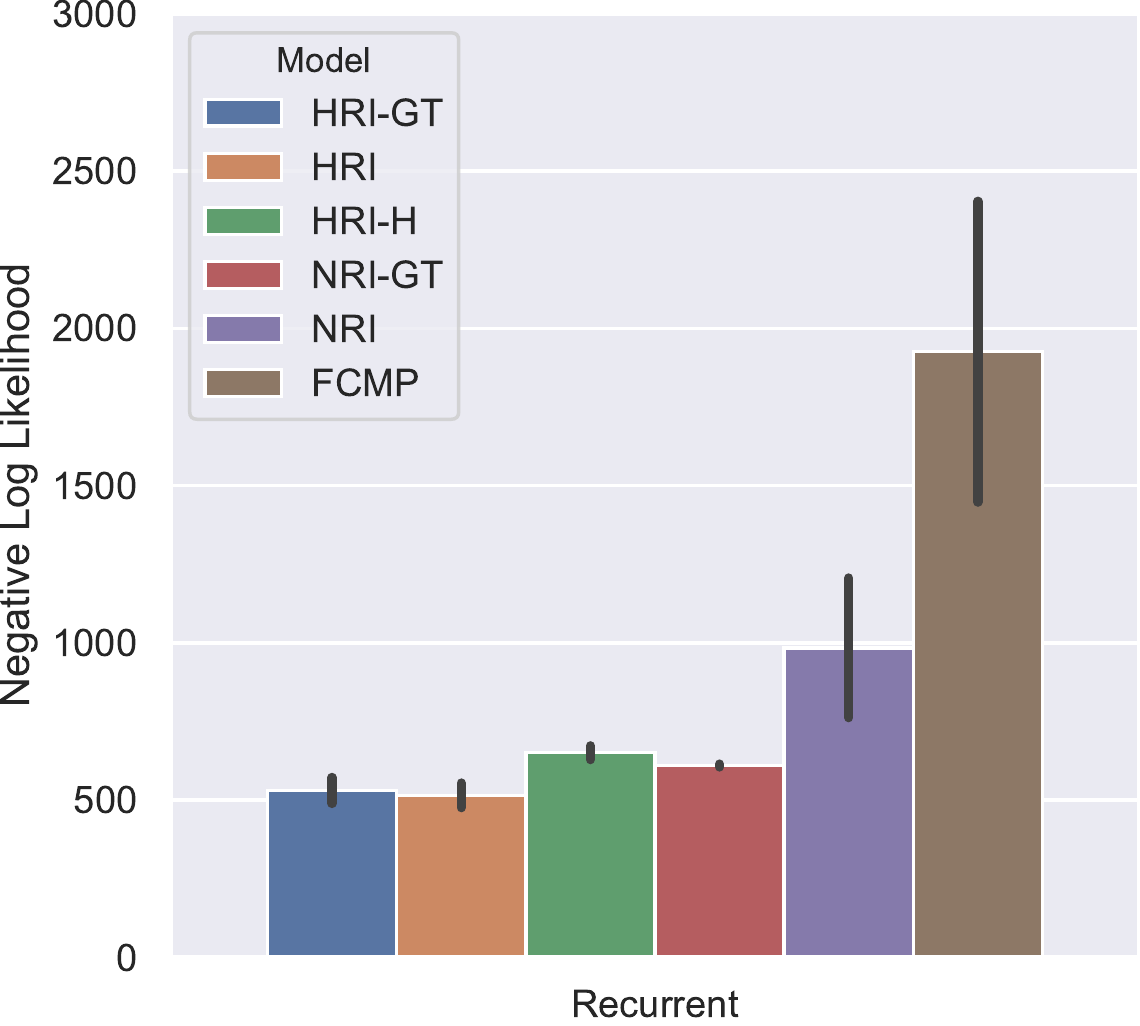}
    \caption{}
    \label{fig:traj-4-3-lstm-nll}
\end{subfigure}
\begin{subfigure}[b]{0.48\columnwidth}   
    \centering 
    \includegraphics[width=\textwidth]{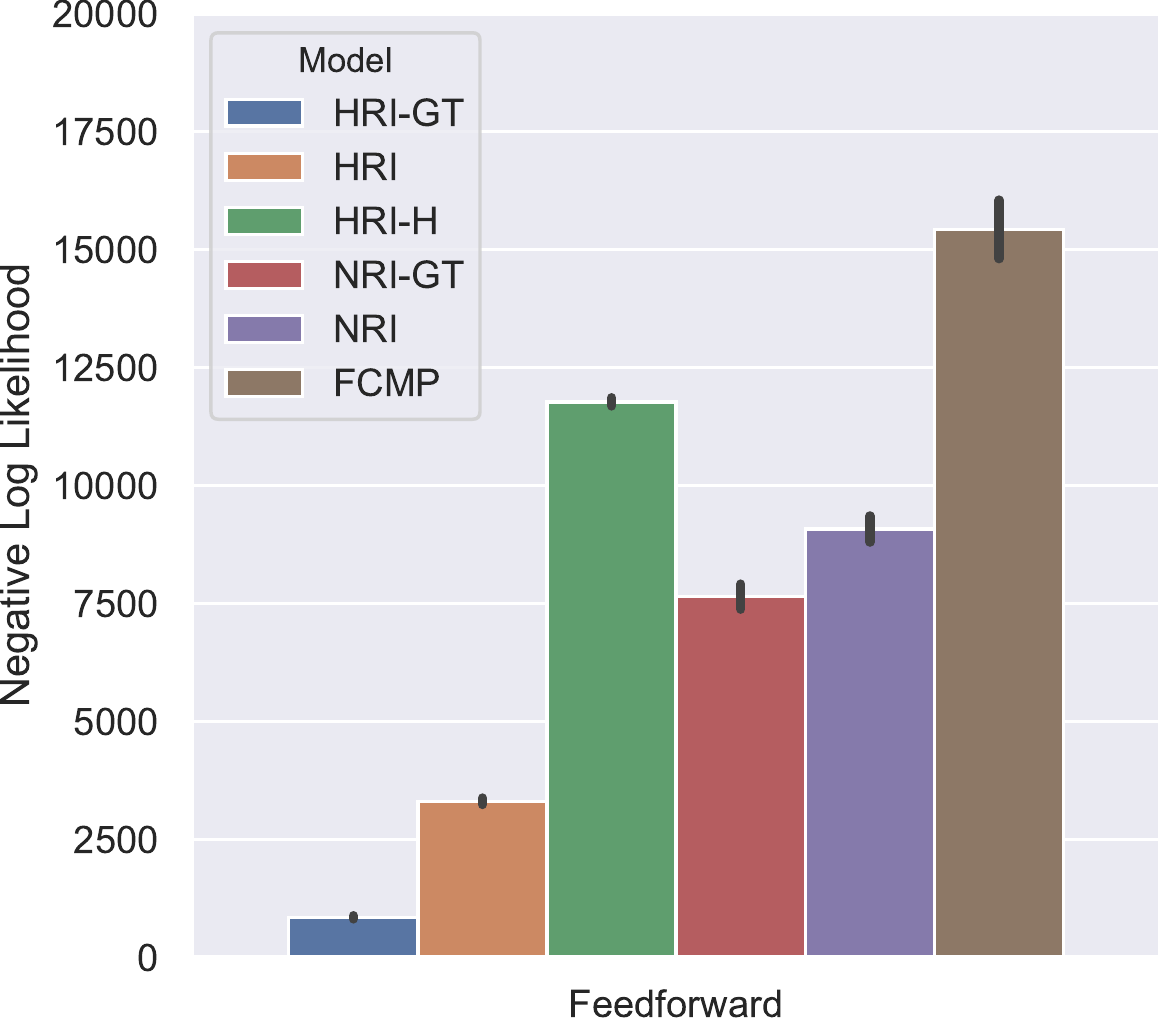}
    \caption{}
    \label{fig:traj-4-3-ff-nll}
\end{subfigure}
\caption{Performance of HRI and ablation models on the 4-3-state-springs dataset in terms of the negative log likelihood in both recurrent and feedforward case. Note the difference in the ranges of y axis indicating the inability of the feedforward model to learn accurate dynamics prediction.   } 
\label{fig:traj-4-3-nll}
\end{figure}

\begin{figure}[h]
\centering
\begin{subfigure}[b]{0.48\columnwidth}   
    \centering 
    \includegraphics[width=\textwidth]{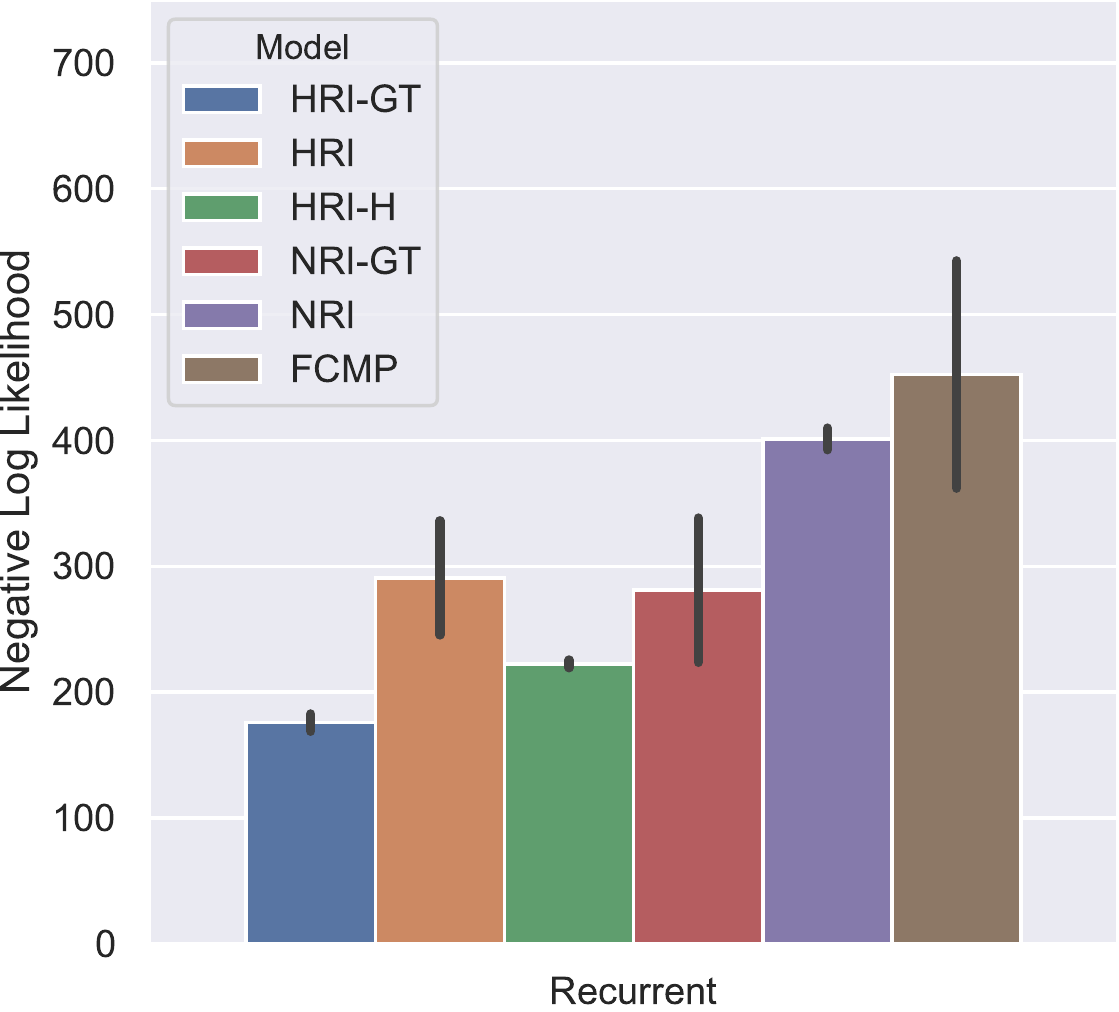}
    \caption{}
    \label{fig:traj-3-3-lstm-nll}
\end{subfigure}
\begin{subfigure}[b]{0.48\columnwidth}   
    \centering 
    \includegraphics[width=\textwidth]{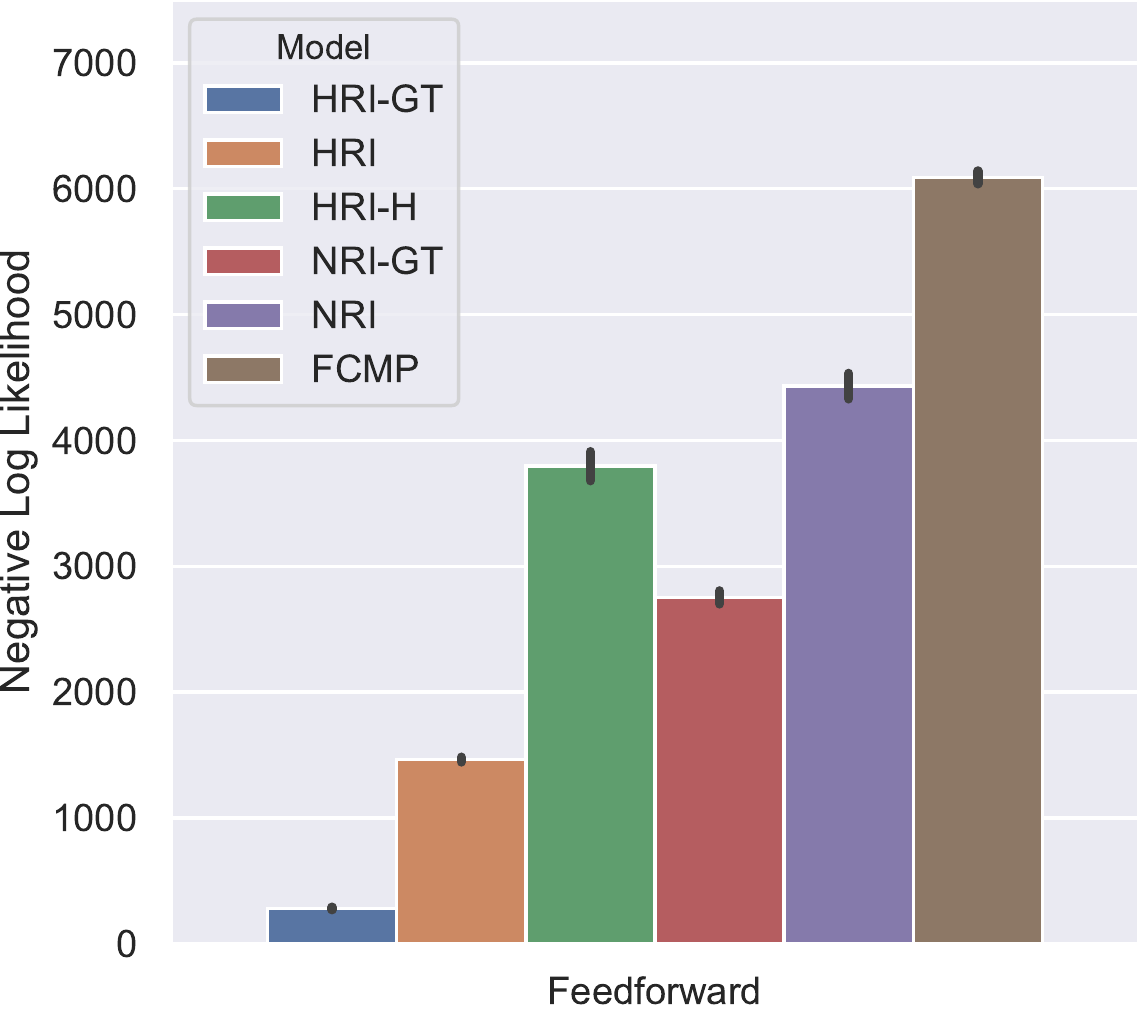}
    \caption{}
    \label{fig:traj-3-3-ff-nll}
\end{subfigure}
\caption{Performance of HRI and ablation models on the 3-3-state-springs dataset in terms of the negative log likelihood in both recurrent and feedforward case. Note the difference in the ranges of y axis indicating the inability of the feedforward model to learn accurate dynamics prediction.   } 
\label{fig:traj-3-3-nll}
\end{figure}

Additional results in terms of the negative log likelihood are provided for the 4-3-state-springs dataset in~\cref{fig:traj-4-3-nll}. 
By having the absolute, instead of the normalized score, we can see how markedly big the gap between the recurrent and the feedforward dynamics predictor is.
Additionally, the results in~\cref{fig:traj-3-3,fig:traj-3-3-nll} for the 3-3-state-springs show that same conclusions hold as for the 4-3-state-springs, with the exception of \cref{fig:traj-3-3-ablation-norm-nll} where HRI and HRI-H perform the same (for 4-3-state-springs HRI outperforms HRI-H, see~\cref{fig:traj-4-3-ablation-norm-nll}).

\onecolumn
\clearpage
\subsection{Visual Springs - Different Number of Objects and Homogeneous Colors}
\label{sec:visual-springs-numbers-bw}

\begin{figure*}[!h]
\centering
\begin{subfigure}[b]{0.3\textwidth}  
    \centering 
    \includegraphics[width=\textwidth]{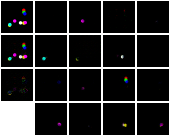}
    \caption{}
    \label{fig:imgs-nll-3-3-slots}
\end{subfigure}
\begin{subfigure}[b]{0.4\textwidth}  
    \centering 
    \includegraphics[width=\textwidth]{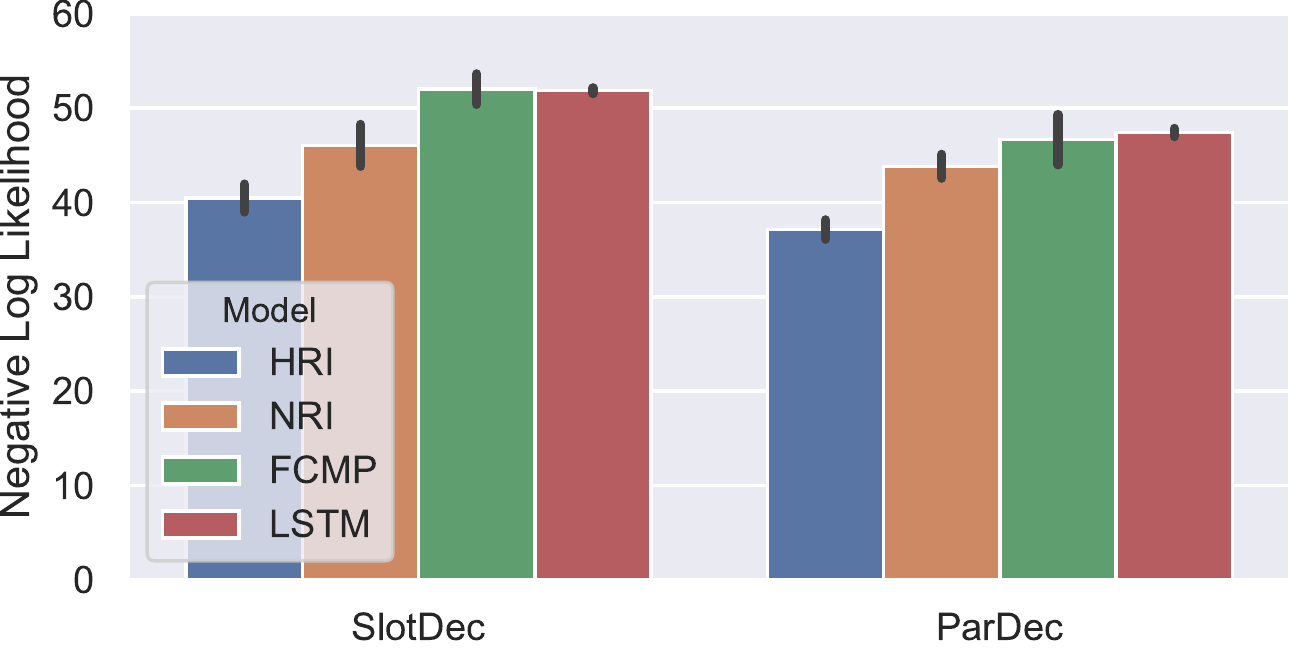}
    \caption{}
    \label{fig:imgs-nll-3-3-bars}
\end{subfigure}
\caption{
(a) HRI introspection: first column contains ground-truth, predicted image and their difference. 
In the other 4 columns we visualize 16 object slots decoded separately. 
(b) Negative log likelihood for all models on the 3-3-visual-springs dataset.
} 
\label{fig:imgs-nll-3-3}
\end{figure*}

\begin{figure*}[!h]
\centering
\begin{subfigure}[b]{0.3\textwidth}  
    \centering 
    \includegraphics[width=\textwidth]{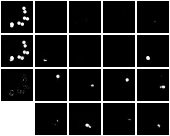}
    \caption{}
    \label{fig:imgs-nll-bw-3-3-slots}
\end{subfigure}
\begin{subfigure}[b]{0.4\textwidth}  
    \centering 
    \includegraphics[width=\textwidth]{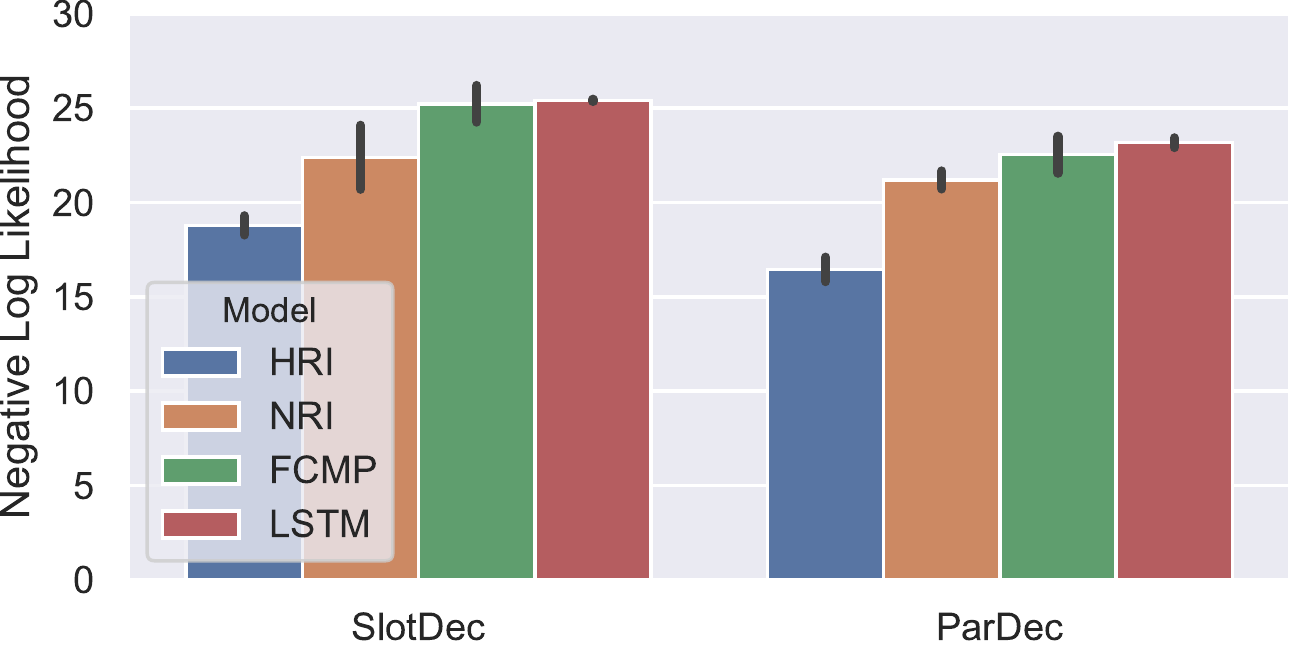}
    \caption{}
    \label{fig:imgs-nll-bw-3-3-bars}
\end{subfigure}
\caption{
(a) HRI introspection: first column contains ground-truth, predicted image and their difference. 
In the other 4 columns we visualize 16 object slots decoded separately. 
(b) Negative log likelihood for all models on 3-3-visual-springs-white.
} 
\label{fig:imgs-nll-bw-3-3}
\end{figure*}

\begin{figure*}[!h]
\centering
\begin{subfigure}[b]{0.3\textwidth}  
    \centering 
    \includegraphics[width=\textwidth]{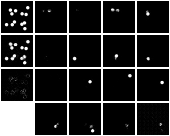}
    \caption{}
    \label{fig:imgs-nll-bw-4-3-slots}
\end{subfigure}
\begin{subfigure}[b]{0.4\textwidth}  
    \centering 
    \includegraphics[width=\textwidth]{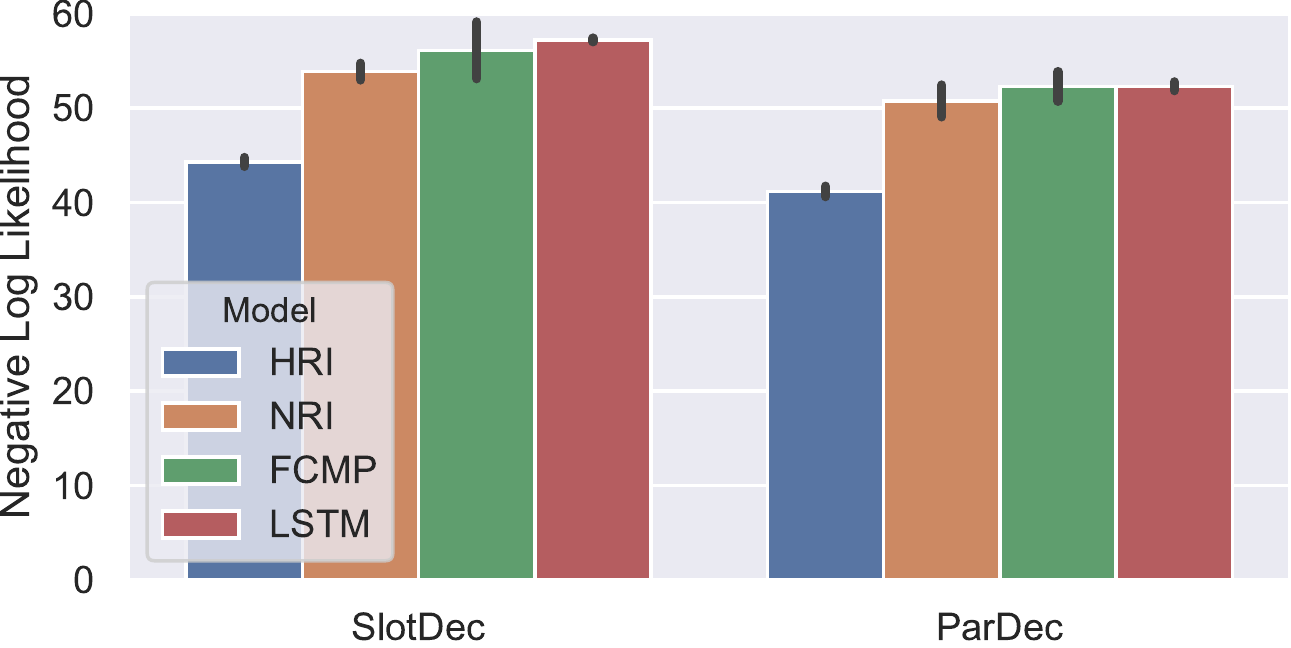}
    \caption{}
    \label{fig:imgs-nll-bw-4-3-bars}
\end{subfigure}
\caption{
(a) HRI introspection: first column contains ground-truth, predicted image and their difference. 
In the other 4 columns we visualize 16 object slots decoded separately. 
(b) Negative log likelihood for all models on 4-3-visual-springs-white.
} 
\label{fig:imgs-nll-bw-4-3}
\end{figure*}

We provide additional results and visualizations for the 3-3-visual-springs and additional two experiments that we performed with homogeneous (white) balls: 3-3-visual-springs-white and 4-3-visual-springs-white.
The result in~\cref{fig:imgs-nll-3-3} confirms that our model is able to handle different number of balls, by learning to simply leave some slots empty (~\cref{fig:imgs-nll-3-3-slots}).
Moreover, the results in~\cref{fig:imgs-nll-bw-3-3,fig:imgs-nll-bw-4-3} closely match the ones in~\cref{fig:imgs-nll-4-3,fig:imgs-nll-3-3}, thus showing that our method does not rely on color to disentangle objects into separate slots.

\subsection{Prediction Rollouts and Latent Interaction Graph Inference}
\label{sec:visual-springs-rollouts-graphs}

\begin{figure*}[!h]
    \centering
    \includegraphics[width=0.8\textwidth]{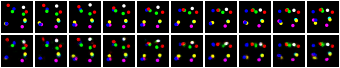}
    \caption{Ground truth (top) and predicted (bottom) 10 time steps rollout of HRI on 4-3-visual-springs.}
    \label{fig:rollout}
\end{figure*}

\begin{figure*}[!h]
\centering 
\includegraphics[width=0.8\textwidth]{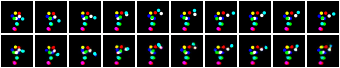}
\caption{ Ground truth (top) and predicted (bottom) 10 time steps rollout of HRI on Human3.6M.}
\label{fig:h36m-rollout}
\end{figure*}

\begin{figure*}[!h]
\centering
\begin{subfigure}[b]{\textwidth}  
    \centering 
    \includegraphics[width=0.8\textwidth]{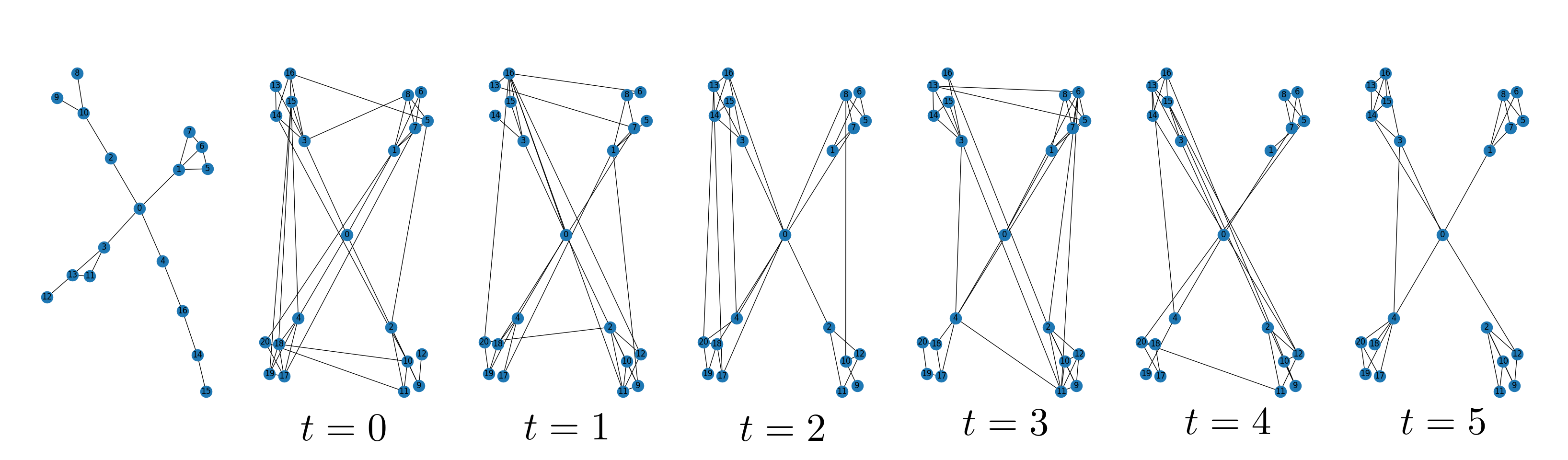}
    \label{fig:inferred-graphs-hri}
\end{subfigure}
\begin{subfigure}[b]{\textwidth}  
    \centering 
    \includegraphics[width=0.8\textwidth]{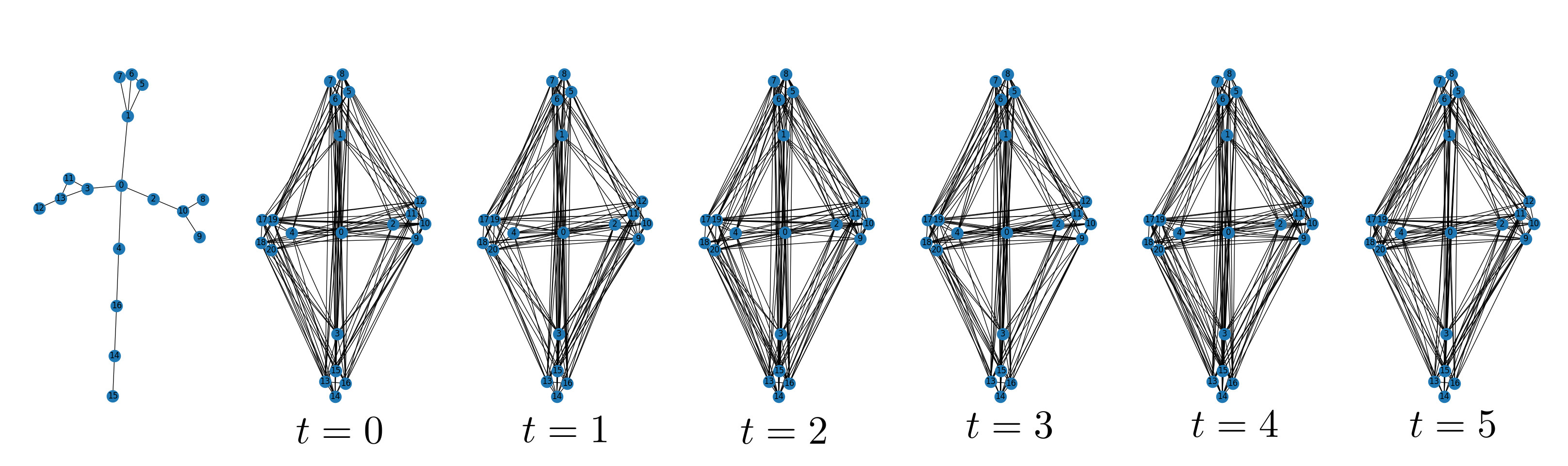}
    \label{fig:inferred-graphs-nri}
\end{subfigure}
\caption{
Inferred interaction graphs on 4-3-visual-springs by HRI (top row) and NRI (bottom row). 
In the first column are groundtruth graphs, and the graphs to follow in subsequent columns correspond to graphs inferred by the respective model over the video sequence frames.
} 
\label{fig:springs-inferred-graphs}
\end{figure*}

We provide additional results for the 4-3-visual-springs and Human3.6M dataset. 
In particular, \cref{fig:rollout} shows 10 time step prediction rollout of HRI model on 4-3-visual springs dataset, where it can be seen that the predictions closely match the ground truth sequence.
The rollout sequence in~\cref{fig:h36m-rollout} shows similar performance on the Human3.6M dataset.
Inferred interaction graphs on 4-3-visual springs by HRI (top row) and NRI (bottom row) are shown in~\cref{fig:springs-inferred-graphs}.
The first column shows the groundtruth graphs, and the graphs to follow in the next columns correspond to graphs inferred by models in different sequence steps.
The graph inferred by HRI resembles the ground-truth much more closely compared to NRI.

\clearpage
\subsection{Visual Springs - Diverse Datasets}
\label{sec:visual-springs-diverse}

\begin{figure}[!h]
\centering
    \begin{subfigure}[b]{0.12\textwidth}  
        \centering 
        \includegraphics[width=\textwidth]{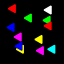}
        \caption{}
        \label{fig:ex-v23}
    \end{subfigure}
\hspace{0.01\textwidth}
    \begin{subfigure}[b]{0.12\textwidth}  
        \centering 
        \includegraphics[width=\textwidth]{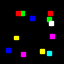}
        \caption{}
        \label{fig:ex-v24}
    \end{subfigure}
\hspace{0.01\textwidth}
    \begin{subfigure}[b]{0.12\textwidth}  
        \centering 
        \includegraphics[width=\textwidth]{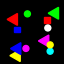}
        \caption{}
        \label{fig:ex-v29}
    \end{subfigure}
\hspace{0.01\textwidth}
    \begin{subfigure}[b]{0.12\textwidth}  
        \centering 
        \includegraphics[width=\textwidth]{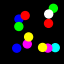}
        \caption{}
        \label{fig:ex-v20-08}
    \end{subfigure}
\hspace{0.01\textwidth}
    \begin{subfigure}[b]{0.12\textwidth}  
        \centering 
        \includegraphics[width=\textwidth]{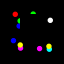}
        \caption{}
        \label{fig:ex-v20c}
    \end{subfigure}
\caption{
Samples from: 
(a) \emph{Triangles}, 
(b) \emph{Squares}, 
(c) \emph{Diverse-Objects},
(d) \emph{Large-Objects}, and 
(e) \emph{Curtain} datasets.
}
\label{fig:diverse-objects-samples}
\end{figure}

\begin{figure}[!h]
\centering
\begin{subfigure}[b]{0.3\textwidth}  
    \centering 
    \includegraphics[width=\textwidth]{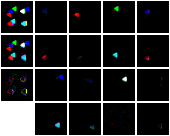}
    \caption{}
    \label{fig:imgs-nll-4-3-slots-v23}
\end{subfigure}
\begin{subfigure}[b]{0.4\textwidth}  
    \centering 
    \includegraphics[width=\textwidth]{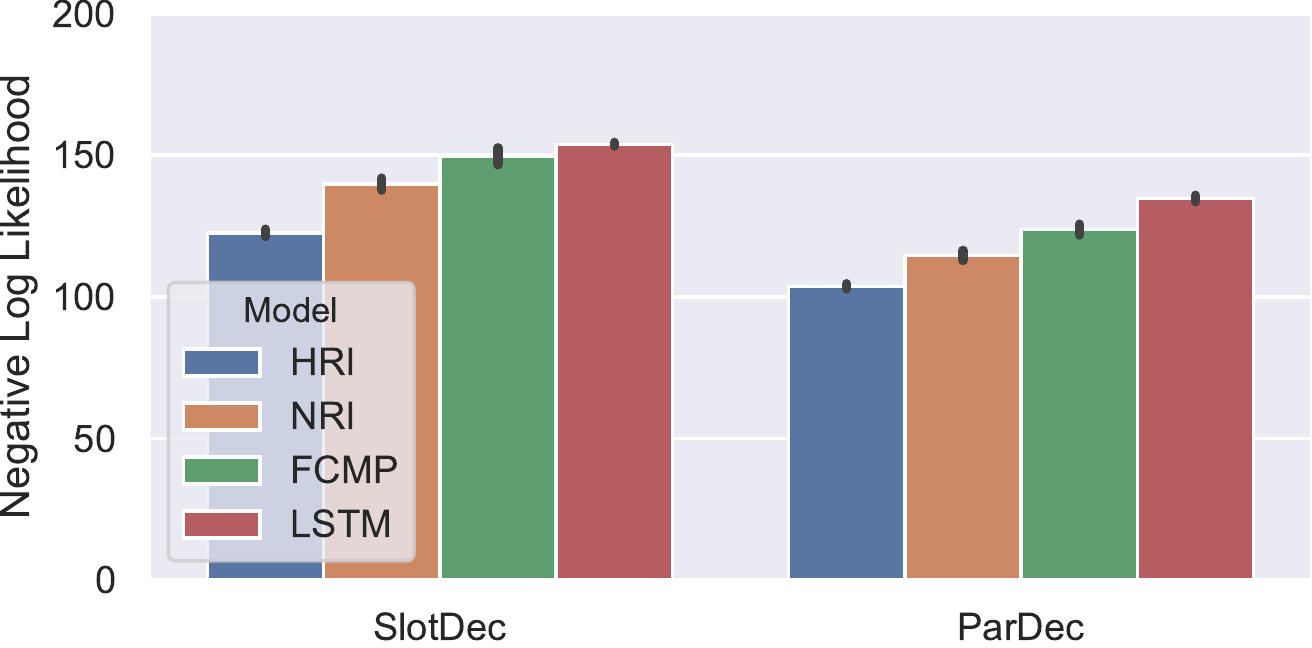}
    \caption{}
    \label{fig:imgs-nll-4-3-par-v23}
\end{subfigure}
\caption{
(a) HRI introspection: first column contains ground-truth, prediction and their difference. The other columns show the 16 object slots decoded separately. 
(b) Negative log likelihood for all models on the \emph{Triangles} dataset.
} 
\label{fig:imgs-nll-4-3-v23}
\end{figure}

\begin{figure}[!h]
\centering
\begin{subfigure}[b]{0.3\textwidth}  
    \centering 
    \includegraphics[width=\textwidth]{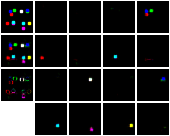}
    \caption{}
    \label{fig:imgs-nll-4-3-slots-v24}
\end{subfigure}
\begin{subfigure}[b]{0.4\textwidth}  
    \centering 
    \includegraphics[width=\textwidth]{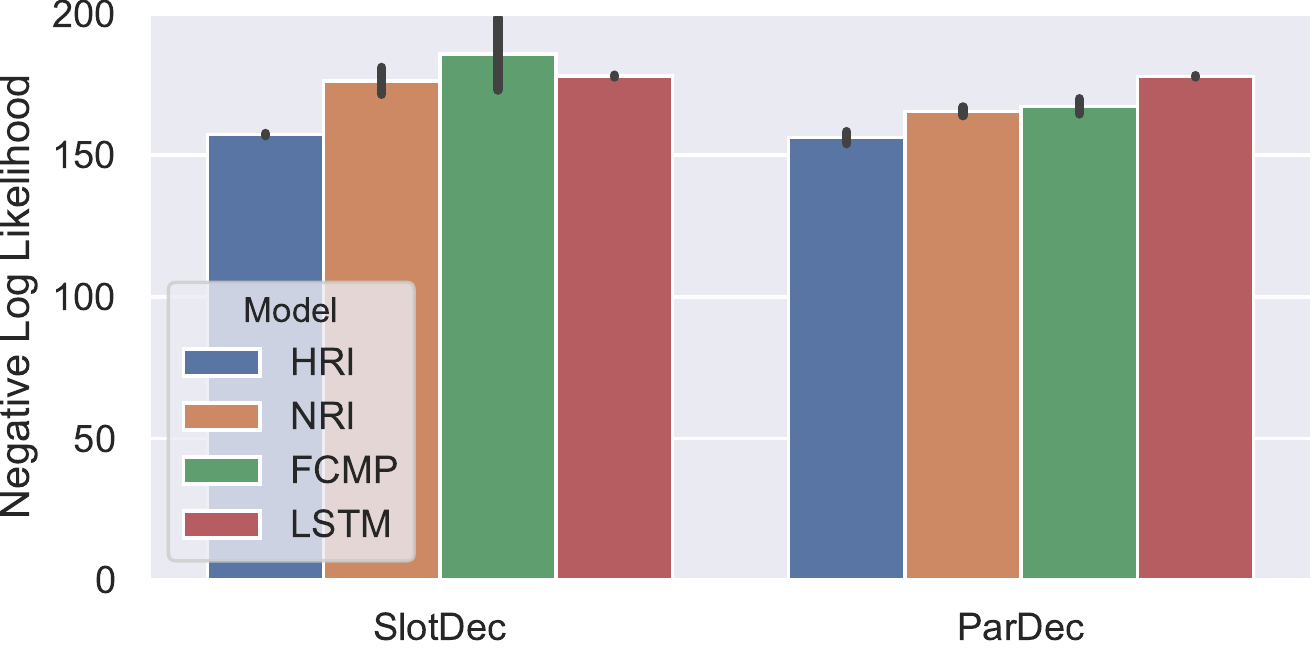}
    \caption{}
    \label{fig:imgs-nll-4-3-par-v24}
\end{subfigure}
\caption{
(a) HRI introspection: first column contains ground-truth, prediction and their difference. The other columns show the 16 object slots decoded separately. 
(b) Negative log likelihood for all models on the \emph{Squares} dataset.
} 
\label{fig:imgs-nll-4-3-v24}
\end{figure}

\begin{figure}[!h]
\centering
\begin{subfigure}[b]{0.3\textwidth}  
    \centering 
    \includegraphics[width=\textwidth]{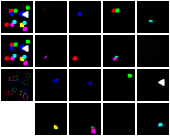}
    \caption{}
    \label{fig:imgs-nll-4-3-slots-v29}
\end{subfigure}
\begin{subfigure}[b]{0.4\textwidth}  
    \centering 
    \includegraphics[width=\textwidth]{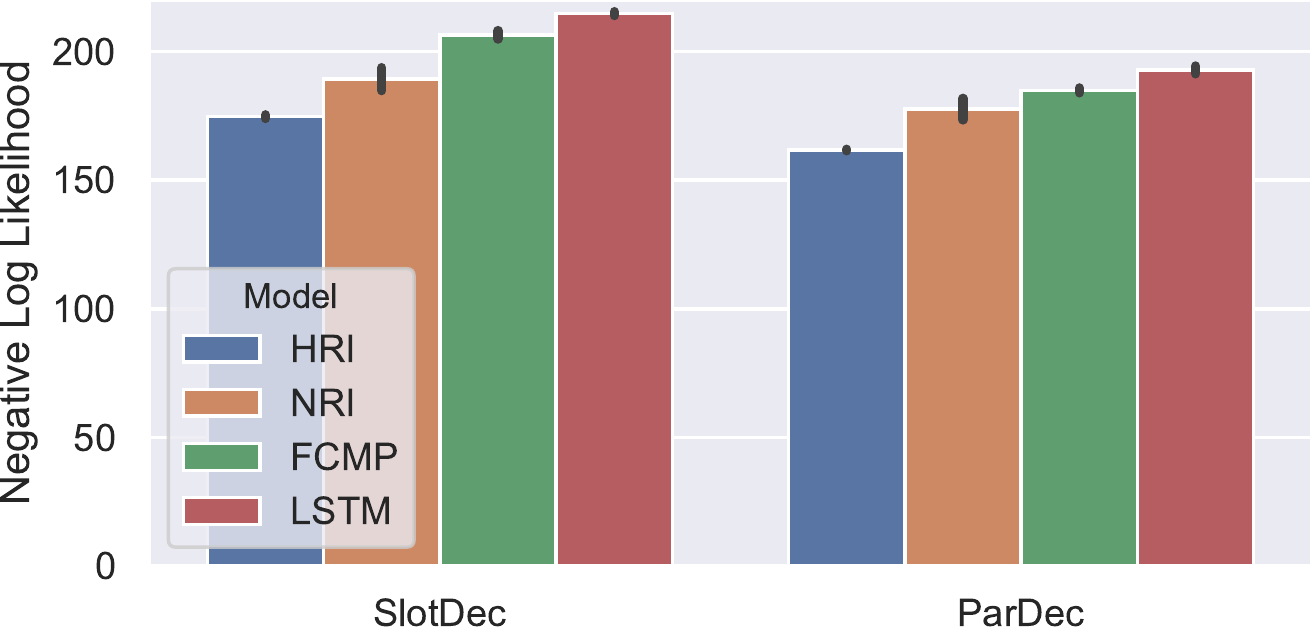}
    \caption{}
    \label{fig:imgs-nll-4-3-par-v29}
\end{subfigure}
\caption{
(a) HRI introspection: first column contains ground-truth, prediction and their difference. The other columns show the 16 object slots decoded separately. 
(b) Negative log likelihood for all models on the \emph{Diverse-objects} dataset.
} 
\label{fig:imgs-nll-4-3-v29}
\end{figure}

\begin{figure}[!h]
\centering
\begin{subfigure}[b]{0.3\textwidth}  
    \centering 
    \includegraphics[width=\textwidth]{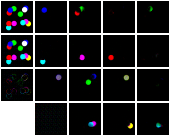}
    \caption{}
    \label{fig:imgs-nll-4-3-slots-v20-08}
\end{subfigure}
\begin{subfigure}[b]{0.4\textwidth}  
    \centering 
    \includegraphics[width=\textwidth]{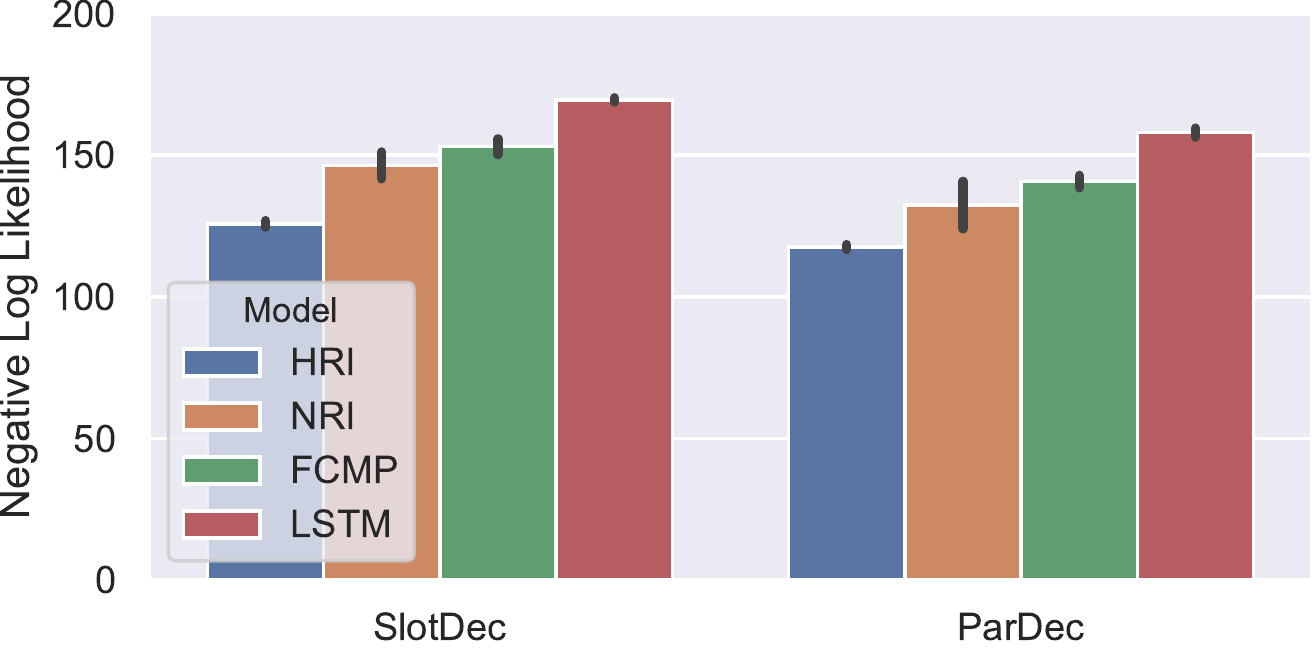}
    \caption{}
    \label{fig:imgs-nll-4-3-par-v20-08}
\end{subfigure}
\caption{
(a) HRI introspection: first column contains ground-truth, prediction and their difference. The other columns show the 16 object slots decoded separately. 
(b) Negative log likelihood for all models on the \emph{Large-Objects} dataset.
} 
\label{fig:imgs-nll-4-3-v20-08}
\end{figure}

\begin{figure}[!h]
\centering
\begin{subfigure}[b]{0.3\textwidth}  
    \centering 
    \includegraphics[width=\textwidth]{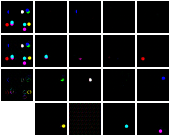}
    \caption{}
    \label{fig:imgs-nll-4-3-slots-v20c}
\end{subfigure}
\begin{subfigure}[b]{0.4\textwidth}  
    \centering 
    \includegraphics[width=\textwidth]{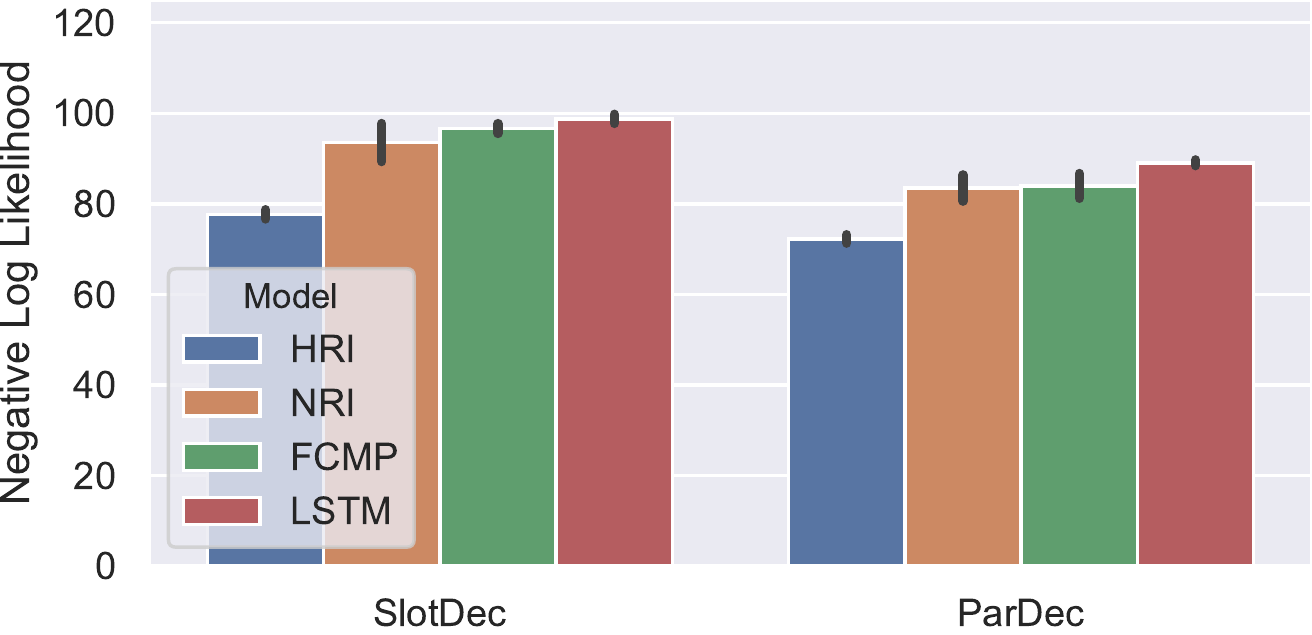}
    \caption{}
    \label{fig:imgs-nll-4-3-par-v20c}
\end{subfigure}
\caption{
(a) HRI introspection: first column contains ground-truth, prediction and their difference. The other columns show the 16 object slots decoded separately. 
(b) Negative log likelihood for all models on the \emph{Curtain} dataset.
} 
\label{fig:imgs-nll-4-3-v20c}
\end{figure}

In this section we provide results on 5 additional datasets: \emph{Triangles}, \emph{Squares}, \emph{Diverse-Objects} (contains balls, triangles and squares of various sizes), \emph{Large-Objects}, and \emph{Curtain}.
Samples of a frame from each of these datasets are shown in~\cref{fig:diverse-objects-samples}.
Qualitative introspection and quantitative performance in~\cref{fig:imgs-nll-4-3-v23,fig:imgs-nll-4-3-v24,fig:imgs-nll-4-3-v29,fig:imgs-nll-4-3-v20-08,fig:imgs-nll-4-3-v20c} show that HRI is able to handle datasets containing different object shapes, sizes, and also handle occlusions, while outperforming all considered baselines in all cases.
\emph{Curtain} dataset is created form 4-3-visual-springs dataset by inserting a $12 \times 12$ pixels curtain at random position in each video sequence.

\clearpage
\subsection{Visual Springs - Generalization to Different Number of Objects}
\label{sec:visual-springs-generalizaton}

\begin{figure}[!h]
\centering
\begin{subfigure}[b]{0.3\textwidth}  
    \centering 
    \includegraphics[width=\textwidth]{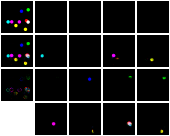}
    \caption{}
    \label{fig:imgs-nll-4-3-gen-to-3-3-slots}
\end{subfigure}
\begin{subfigure}[b]{0.4\textwidth}  
    \centering 
    \includegraphics[width=\textwidth]{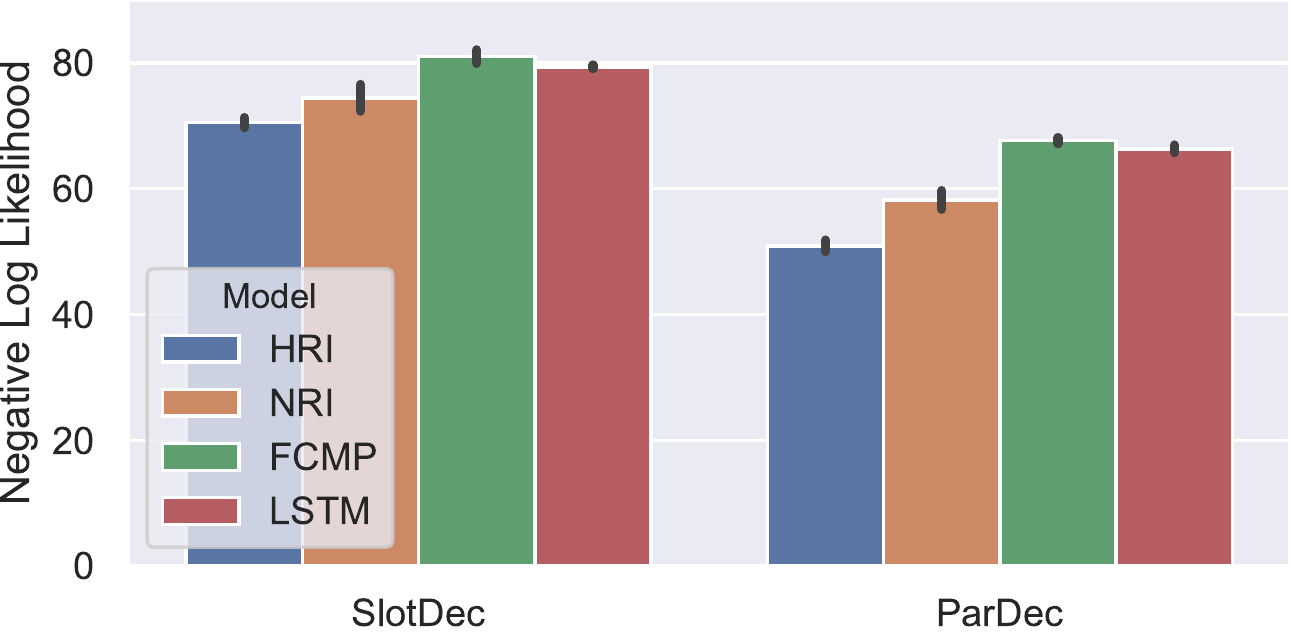}
    \caption{}
    \label{fig:imgs-nll-4-3-gen-to-3-3-bars}
\end{subfigure}
\caption{
(a) HRI introspection: first column contains ground-truth, prediction and their difference. The other columns show the 16 object slots decoded separately. 
(b) Negative log likelihood for all models on trained on 4-3-visual-springs and evaluated on 3-3-visual-springs dataset.
} 
\label{fig:imgs-nll-4-3-gen-to-3-3}
\end{figure}

\begin{figure}[!h]
    \centering
    \includegraphics[width=0.8\textwidth]{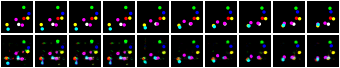}
    \caption{Ground truth (top) and predicted (bottom) 10 time steps rollout of HRI trained on 4-3-visual springs and evaluated on 3-3-visual-springs.}
    \label{fig:gen-4-3-to-3-3-rollout}
\end{figure}

We test whether HRI is able to generalize to scenarios with different number of objects by first training on 4-3-visual-springs and then testing on 3-3-visual springs.
The qualitative results are shown in~\cref{fig:imgs-nll-4-3-gen-to-3-3-slots}, where it can be seen that HRI is able to infer separate object representations even on a dataset it was not originally trained on.
Quantitative comparison of model performance is shown in~\cref{fig:imgs-nll-4-3-gen-to-3-3-bars}.
Compared to~\cref{fig:imgs-nll-3-3-bars} we observe a slight drop in performance, but HRI remains the best performing model.
A prediction sequence is shown in~\cref{fig:gen-4-3-to-3-3-rollout}.
Note how HRI makes minor mistakes in the first few frames, but adapts afterwards to the setting with different number of objects from what it was trained on, and models the dynamics more accurately in subsequent frames.

\end{document}